\documentclass[review,11pt]{ReportTemplate}
\usepackage{bm}
\usepackage{graphicx}
\usepackage{paralist,algorithmic,algorithm}
\usepackage{amsmath,mathrsfs,amssymb}

\newtheorem{theorem}{Theorem}
\newtheorem{definition}{Definition}
\newtheorem{lemma}{Lemma}

\begin{document}

\begin{frontmatter}
  \title{Theoretical Exploration of Flexible Transmitter Model}

  \author{Jin-Hui Wu}
  \author{Shao-Qun Zhang}
  \author{Yuan Jiang}
  \author{Zhi-Hua Zhou\footnote{Zhi-Hua Zhou is the corresponding author.}}

  \address{National Key Laboratory for Novel Software Technology\\
  Nanjing University, Nanjing 210093, China\\
  \{wujh,zhangsq,jiangy,zhouzh\}@lamda.nju.edu.cn}

  \begin{abstract}
    Neural network models generally involve two important components, i.e., network architecture and neuron model. Although there are abundant studies about network architectures, only a few neuron models have been developed, such as the MP neuron model developed in 1943 and the spiking neuron model developed in the 1950s. Recently, a new bio-plausible neuron model, Flexible Transmitter (FT) model~\cite{zhang2021flexible}, has been proposed. It exhibits promising behaviors, particularly on temporal-spatial signals, even when simply embedded into the common feedforward network architecture. This paper attempts to understand the properties of the FT network (FTNet) theoretically. Under mild assumptions, we show that: i) $\mathrm{FTNet}$ is a universal approximator; ii) the approximation complexity of $\mathrm{FTNet}$ can be exponentially smaller than those of commonly-used real-valued neural networks with feedforward/recurrent architectures and is of the same order in the worst case; iii) any local minimum of $\mathrm{FTNet}$ is the global minimum, implying that it is possible to identify global minima by local search algorithms. 
  \end{abstract}

\end{frontmatter}

\section{Introduction}

Deep neural networks have become mainstream in artificial intelligence and have exhibited excellent performance in many applications, such as disease detection~\cite{laguarta2021longitudinal}, machine translation~\cite{bahdanau2015neural}, emotion recognition~\cite{guo2021singular}, etc. Typically, a neural network model is composed of a network architecture and a neuron model. The past decade has witnessed abundant studies about network architectures, whereas the modeling of neurons is relatively less considered. Typical neuron models include the MP neuron model~\cite{mcculloch1943logical} and the spiking neuron model~\cite{hodgkin1952quantitative,zhang2021bifurcation}. Recently, a new bio-plausible neuron model, \emph{Flexible Transmitter} model~\cite{zhang2021flexible}, has been proposed. In contrast to the classical neuron models, the FT neuron model mimics neurotrophic potentiation and depression effects by a formulation of a two-variable function, exhibiting great potential for temporal-spatial data processing. Furthermore, Zhang and Zhou~\cite{zhang2021flexible} developed the Flexible Transmitter Network, a feed-forward neural network composed of FT neurons, which performs competitively with state-of-the-art models when handling temporal-spatial data. 

However, the theoretical properties of the FT model remain unknown. This work takes one step in this direction. We notice that the formulation of the FT model provides greater flexibility for the representation of neuron models, and its benefits are twofold. Firstly, the complex-valued implementation takes into account the magnitude and phase of variables and is thus good at processing data with norm-preserving and antisymmetric structures. Secondly, the modeling of neurotrophic potentiation and depression effects derives a local recurrent system, and $\mathrm{FTNet}$ intrinsically has temporal-spatial representation ability even in a feed-forward architecture. Inspired by these insights, we present the theoretical advantages over the Feedforward Neural Network ($\mathrm{FNN}$) and Recurrent Neural Network ($\mathrm{RNN}$) from the perspectives of approximation and local minima. Our main contributions can be summarized as follows: 	
\begin{enumerate}
    \item $\mathrm{FTNet}$ is a universal approximator, i.e., a one-hidden-layer $\mathrm{FTNet}$ with admissible activation functions can approximate any continuous function and discrete-time open dynamical system on any compact set arbitrarily well, stated in Theorems~\ref{thm: ua of f-ftnet} and \ref{thm: ua of r-ftnet}, respectively. 
    \item We present the approximation-complexity advantages and the worst-case guarantees of $\mathrm{FTNet}$ over the $\mathrm{FNN}$ and $\mathrm{RNN}$. Specifically, separation results exist between one-hidden-layer $\mathrm{FTNet}$ and one-hidden-layer $\mathrm{FNN}$/$\mathrm{RNN}$, as shown in Theorems~\ref{thm: advantage of ftnet over fnn}~and~\ref{thm: advantage of ftnet over rnn}, respectively. In addition, any function expressible by a one-hidden-layer $\mathrm{FNN}$ or $\mathrm{RNN}$ can be approximated by a one-hidden-layer $\mathrm{FTNet}$ with a similar number of hidden neurons, as shown in Theorems~\ref{thm: worst case guarantee of ftnet over fnn}~and~\ref{thm: worst case guarantee of ftnet over rnn}, respectively. These theorems imply that $\mathrm{FTNet}$ is capable of expressing functions more efficiently than $\mathrm{FNN}$ and $\mathrm{RNN}$. 
    \item We show that $\mathrm{FTNet}$ in the feedforward architecture has no suboptimal local minimum using general activations and loss functions, as illustrated in Theorem~\ref{thm: local minima of f-ftnet has loss 0}. This implies that local search algorithms for $\mathrm{FTNet}$ have the potential to converge to the global minimum. 
\end{enumerate}

The rest of this paper is organized as follows. Section~\ref{sec: related work} introduces related work. Section~\ref{sec: preliminary} provides basic notations, definitions, and the formulation of $\mathrm{FTNet}$. Section~\ref{sec: universal approximation} proves the universal approximation of $\mathrm{FTNet}$. Section~\ref{sec: approximation complexity} investigates the approximation complexity of $\mathrm{FTNet}$. Section~\ref{sec: local minima} studies the property of the local minima of $\mathrm{FTNet}$. Section~\ref{sec: conclusion and prospect} concludes our work with prospect.

\section{Related Works}
\label{sec: related work}

\noindent \textbf{Universal Approximation.} The universal approximation confirms the powerful expressivity of neural networks. The earliest research is the universal approximation theorem of $\mathrm{FNN}$, which proves that $\mathrm{FNN}$ with suitable activation functions can approximate any continuous function on any compact set arbitrarily well~\cite{cybenko1989approximation,funahashi1989approximate,hornik1989multilayer}. Furthermore, Leshno et al. point out that a non-polynomial activation function is the necessary and sufficient condition for $\mathrm{FNNs}$ to achieve universal approximation~\cite{leshno1993multilayer}. Later, some researchers extend the universal approximation theorems to other real-weighted neural networks with different architectures, such as $\mathrm{RNN}$~\cite{seidl1991structure,funahashi1993approximation,chow2000modeling,li2005approximation,schafer2006recurrent} and convolutional neural networks~\cite{zhou2020universality}. For complex-weighted neural networks, it has been proven that they can approximate any continuous complex-valued functions on any compact set using some activation function~\cite{arena1993capability}, and that non-holomorphic and non-antiholomorphic activation functions are the necessary and sufficient condition of universal approximation~\cite{voigtlaender2020universal}. Our work investigates the universal approximation of $\mathrm{FTNet}$, i.e., the capability of complex-weighted neural networks to approximate real-valued functions and dynamical systems. 

\noindent \textbf{Approximation Complexity.} The universal approximation theorems only prove the possibility of approximating certain functions, but do not consider approximation complexity, i.e., the number of required hidden neurons for approximating particular functions. It is also important to consider approximation complexity, which reflects the efficiency of approximation. Early works focus on the degree of approximation of one-hidden-layer $\mathrm{FNN}$, i.e., how approximation error depends on the input dimension and the number of hidden units~\cite{barron1993universal,debao1993degree,hornik1994degree,mhaskar1994dimension}. Recent works prove separation results, i.e., one model cannot be expressed by another model with the same order of parameters~\cite{eldan2016power,safran2017depth,telgarsky2016benefits,lu2017expressive}. A notable work proves that one-hidden-layer $\mathrm{FNN}$ needs at least exponential parameters to express a given complex-reaction network~\cite{zhang2022towards}. Our work not only provides the separation results between $\mathrm{FTNet}$ and $\mathrm{FNN}$/$\mathrm{RNN}$ but also guarantees that any $\mathrm{FNN}$/$\mathrm{RNN}$ can be expressed by $\mathrm{FTNet}$ with a similar hidden size. 

\noindent \textbf{Local Minima.} Suboptimal local minima are undesirable points of the loss surface, without which it is tractable to train neural networks using local search algorithms. Early works show that one-hidden-layer $\mathrm{FNN}$ using the squared loss has no suboptimal local minimum under suitable conditions~\cite{poston1991local,yu1992can,yu1995local,huang1998local}. These results are extended to multilayer $\mathrm{FNN}$~\cite{nguyen2017loss,kawaguchi2020elimination} and other types of neural networks, such as deep ResNet~\cite{kawaguchi2019depth}, deep convolutional neural networks~\cite{nguyen2018optimization}, deep linear networks~\cite{kawaguchi2016deep,laurent2018deep}, and over-parameterized deep neural networks~\cite{nguyen2018loss}. From another perspective of algorithms, some researchers prove that some commonly used gradient-based algorithms, e.g., GD and SGD, can converge to the global minimum or an almost optimal solution when optimizing over-parameterized neural networks~\cite{gori1992problem,du2018gradient,jacot2018neural,li2018learning,allen2019convergence,du2019gradient,allen2019convergence2,zou2019improved,zou2020gradient,liu2022loss,zhao2021convergence}. Our work extends the classical results of $\mathrm{FNN}$ to $\mathrm{FTNet}$ in the feedforward architecture and generalizes the condition on the loss function from the squared loss to a large class of analytic functions.

\section{Preliminary}
\label{sec: preliminary}

We denote by $\mathrm{i} = \sqrt{-1}$ the imaginary unit. Let $\mathrm{Re}(z)$, $\mathrm{Im}(z)$, $\theta_z$, and $\overline{z}$ be the real part, imaginary part, phase, and complex conjugate of the complex number $z$, respectively. Let $\mathbf{0}^{a \times b}$ denote the zero matrix with $a$ rows and $b$ columns. 

This work considers $\mathrm{FTNet}$ with two typical architectures, that is, Recurrent $\mathrm{FTNet}$ ($\mathrm{R}\text{-}\mathrm{FTNet}$) and Feedforward $\mathrm{FTNet}$ ($\mathrm{F}\text{-}\mathrm{FTNet}$), and the time series regression task with $1$-dimensional outputs throughout this paper. We focus on one-hidden-layer $\mathrm{FTNet}$ throughout this paper. For deep $\mathrm{FTNet}$, it would be interesting to study feature space transformation, which might be a key to understanding the mysteries behind the success of deep neural networks~\cite{zhou2021over}. Let $\bm{x}_t \in \mathbb{R}^I$ be the input vector at time $t$, and $\bm{x}_{1:T} = ( \bm{x}_1 ; \bm{x}_2 ; \dots ; \bm{x}_T ) \in \mathbb{R}^{IT}$ denotes the concatenated input vector at time $T$. We employ the mapping $f_{\times,\mathrm{R}}$ to denote a one-hidden-layer $\mathrm{R}\text{-}\mathrm{FTNet}$ with $H_\times \geqslant I+1$ hidden neurons as follows
\begin{equation}
    \label{eq: mapping of r-ftnet}
    \begin{aligned}
        &~ f_{\times,\mathrm{R}} : \bm{x}_{1:T} \mapsto (y_{\times,1},\dots,y_{\times,T}) , \\ 
        &~ \bm{s}_t + \bm{r}_t \mathrm{i} = \sigma_{\mathrm{\times}} ( \left( \mathbf{W}_{\mathrm{\times}} + \mathbf{V}_{\mathrm{\times}} \mathrm{i} \right) \left( \kappa(\bm{x}_t , H_\times) + \bm{r}_{t-1} \mathrm{i} \right) ) , \\ 
        &~ y_{\times,t} = \bm{\alpha}_{\mathrm{\times}}^\top \bm{s}_t , \quad t \in [T] , 
    \end{aligned}
\end{equation}
where $\bm{r}_t$, $\bm{s}_t \in \mathbb{R}^{H_\times}$, and $y_{\times,t} \in \mathbb{R}$ represent the receptor, stimulus, and output at time $t$, respectively, $\mathbf{W}_\times$, $\mathbf{V}_\times$, $\bm{\alpha}_\times$ denote real-valued weight parameters, $\kappa : \mathbb{R}^I \times \mathbb{N}^+ \rightarrow \mathbb{R}^{H_\times}$ stretches the input to a higher-dimensional space in which
\begin{equation}
	\label{eq: definition of phi}
	\kappa(\bm{x},H_\times) = (\bm{x};0;\dots;0;1) \in \mathbb{R}^{H_\times} \quad \mathrm{with} \quad H_\times \geqslant I+1 ,
\end{equation}
and $\sigma_\times$ is an activation function applied componentwise. Notice that Eq.~\eqref{eq: mapping of r-ftnet} is a multiplicative (rather than additive) form of $\mathrm{FTNet}$ since multiplication is the last operation before applying the activation function. In addition, we also employ the mapping $f_{\times,\mathrm{F}}$ to denote a one-hidden-layer $\mathrm{F}\text{-}\mathrm{FTNet}$ with $H_\times \geqslant I+1$ hidden neurons as follows
\begin{equation}
    \label{eq: mapping of f-ftnet}
    f_{\times,\mathrm{F}} : \bm{x} \mapsto \bm{\alpha}_\times^\top \mathrm{Re} \left[ \sigma_\times \left( \left( \mathbf{W}_{\mathrm{\times}} + \mathbf{V}_{\mathrm{\times}} \mathrm{i} \right) \kappa(\bm{x} , H_\times) \right) \right] . 
\end{equation}

The $\mathrm{zReLU}$ activation function~\cite{guberman2016complex} is a promising choice of the activation function in $\mathrm{FTNet}$, which extends the widely used real-valued activation function $\mathrm{ReLU}$~\cite{fukushima1969visual} to the complex-valued domain, and is defined as 
\begin{equation}
    \label{eq: definition of zrelu}
    \sigma(z) = \left\{ 
        \begin{array}{ll}
            z , & \mathrm{if} \quad \theta_z \in \left[ 0 , \pi / 2 \right] \cup \left[ \pi , 3\pi / 2 \right] , \\ 
            0 , & \mathrm{otherwise} . 
        \end{array}
    \right. 
\end{equation}

Dynamical systems are of great interest when considering the universal approximation of neuron models in the recurrent architecture. We focus on the discrete-time open dynamical system defined as follows. 
\begin{definition}
    Given an initial hidden state $\bm{h}_0 \in \mathbb{R}^{H_\mathrm{D}}$ with $H_\mathrm{D} \in \mathbb{N}^+$, a Discrete-time Open Dynamical System ($\mathrm{DODS}$) is a mapping $f_\mathrm{D}$ defined by 
    \begin{equation}
        \label{eq: mapping of dods}
        \begin{aligned}
        &~ f_\mathrm{D} : \bm{x}_{1:T} \mapsto \left( y_1 , \cdots , y_T \right) , \\ 
        &~ y_t = \psi ( \bm{h}_t ) , \\
        &~ \bm{h}_t = \varphi ( \bm{x}_t , \bm{h}_{t-1} ) , 
        \quad 
        t \in [T] , 
        \end{aligned}
    \end{equation}
    where $\bm{x}_t \in \mathbb{R}^I$, $\bm{h}_t \in \mathbb{R}^{H_\mathrm{D}}$, and $y_t \in \mathbb{R}$ represent the input, hidden state, and output at time $t$, respectively, $\varphi : \mathbb{R}^I \times \mathbb{R}^{H_\mathrm{D}} \rightarrow \mathbb{R}^{H_\mathrm{D}}$ and $\psi : \mathbb{R}^{H_\mathrm{D}} \rightarrow \mathbb{R}^O$ are continuous mappings. 
\end{definition}

\section{Universal Approximation}
\label{sec: universal approximation}

We show the universal approximation of $\mathrm{F}\text{-}\mathrm{FTNet}$ and $\mathrm{R}\text{-}\mathrm{FTNet}$ in Subsections~\ref{subsec: universal approximation of f-ftnet} and~\ref{subsec: universal approximation of r-ftnet}, respectively. 

\subsection{Universal Approximation of \emph{F-FTNet}}
\label{subsec: universal approximation of f-ftnet}

Let $\Vert f \Vert_{L^\infty(\Omega)}$ denote the essential supremum of the function $f$ on the domain $\Omega$, i.e., 
\begin{equation*}
    \Vert f \Vert_{L^\infty(\Omega)} = \inf \left\{ \lambda \mid \mu \left\{ x : |f(x)| \geqslant \lambda \right\} = 0 \right\} , 
\end{equation*}where $\mu$ is the Lebesgue measure. We now present the universal approximation for $\mathrm{F}\text{-}\mathrm{FTNet}$ as follows. 

\begin{theorem}
    \label{thm: ua of f-ftnet}
    Let $K \subset \mathbb{R}^I$ be a compact set, $g$ is a continuous function on $K$, and $\sigma_\times$ is the activation function of $\mathrm{F}\text{-}\mathrm{FTNet}$. Suppose there exists a constant $c \in \mathbb{R}$, such that the function $\sigma(x) = \mathrm{Re} \left[ \sigma_\times(x+c\mathrm{i}) \right]$ is continuous almost everywhere and not polynomial almost everywhere. Then for any $\varepsilon > 0$, there exists an $\mathrm{F}\text{-}\mathrm{FTNet}$ $f_{\times,\mathrm{F}}$, such that 
    \begin{equation*}
        \left\Vert f_{\times,\mathrm{F}} - g \right\Vert_{L^\infty(K)} 
        \leqslant \varepsilon . 
    \end{equation*}
\end{theorem}

Theorem~\ref{thm: ua of f-ftnet} indicates that $\mathrm{F}\text{-}\mathrm{FTNet}$ with suitable activation functions can approximate any continuous function on any compact set arbitrarily well. The conditions in this theorem are satisfied by many commonly used activation functions, such as $\mathrm{modReLU}$~\cite{arjovsky2016unitary}, $\mathrm{zReLU}$~\cite{guberman2016complex}, and $\mathbb{C}\mathrm{ReLU}$~\cite{trabelsi2018deep}. Previous studies focus on the universal approximation of real-weighted networks with real-valued target functions or complex-weighted networks with complex-valued target functions. To our knowledge, Theorem~\ref{thm: ua of f-ftnet} is the first result considering the approximation capability of complex-weighted networks with real-valued target functions. The condition about $\sigma$ of $\mathrm{FTNet}$ is the same as that of $\mathrm{FNN}$~\cite{leshno1993multilayer}, but the activation $\sigma_\times$ of $\mathrm{FTNet}$ is more flexible since $\sigma$ is just the restriction of $\sigma_\times$ on a particular direction. The requirement of not polynomial $\sigma$ is weaker than non-holomorphic and non-antiholomorphic activation, which is the necessary requirement of complex-weighted networks with complex-valued target functions~\cite{voigtlaender2020universal}. Thus, $\mathrm{FTNet}$ successfully benefits from expressing real-valued functions instead of complex-valued ones. We begin our proof with the following lemma.

\begin{lemma}{\cite[Theorem 1]{leshno1993multilayer}}
    \label{lem: ua of fnn}
    Let $K \subset \mathbb{R}^I$ be a compact set, and $g$ is a continuous function on $K$. Suppose the activation function $\sigma_\mathrm{F}$ is continuous almost everywhere and not polynomial almost everywhere. Then for any $\varepsilon > 0$, there exist $H_\mathrm{F} \in \mathbb{N}^+$, $\mathbf{W}_\mathrm{F} \in \mathbb{R}^{H_\mathrm{F} \times I}$, and $\bm{\theta}_\mathrm{F} , \bm{\alpha}_\mathrm{F} \in \mathbb{R}^{H_\mathrm{F}}$, such that 
    \begin{equation*}
        \left\Vert \bm{\alpha}_\mathrm{F}^\top \sigma_\mathrm{F} \left( \mathbf{W}_\mathrm{F} \bm{x} - \bm{\theta}_\mathrm{F} \right) - g(\bm{x}) \right\Vert_{L^\infty(K)} 
        \leqslant \varepsilon . 
    \end{equation*}
\end{lemma}

Lemma~\ref{lem: ua of fnn} shows that one-hidden-layer $\mathrm{FNN}$ can approximate any continuous function on any compact set arbitrarily well, using suitable activation functions. 

\noindent \textbf{Proof of Theorem~\ref{thm: ua of f-ftnet}.} Based on Lemma~\ref{lem: ua of fnn}, it suffices to construct an $\mathrm{F}\text{-}\mathrm{FTNet}$ that has the same output as any given $\mathrm{FNN}$. Since the function $\sigma(x)$ is continuous almost everywhere and not polynomial almost everywhere, it satisfies the conditions in Lemma~\ref{lem: ua of fnn}. According to Lemma~\ref{lem: ua of fnn}, there exist $H_\mathrm{F} \in \mathbb{N}^+$, $\mathbf{W}_\mathrm{F} \in \mathbb{R}^{H_\mathrm{F} \times I}$, and $\bm{\theta}_\mathrm{F} , \bm{\alpha}_\mathrm{F} \in \mathbb{R}^{H_\mathrm{F}}$, such that 
\begin{equation}
    \label{eq: in the proof of ua of f-ftnet, ua of fnn}
    \left\Vert \bm{\alpha}_\mathrm{F}^\top \sigma \left( \mathbf{W}_\mathrm{F} \bm{x} - \bm{\theta}_\mathrm{F} \right) - g(\bm{x}) \right\Vert_{L^\infty(K)} 
    \leqslant \varepsilon . 
\end{equation}
We now construct an $\mathrm{F}\text{-}\mathrm{FTNet}$ with $H_\times = \max \{ I+1 , H_\mathrm{F} \}$ hidden neurons as follows 
\begin{equation*}
    \mathbf{W}_\times = \left[ 
        \begin{array}{ccc}
            \mathbf{W}_\mathrm{F} & \bm{0} & - \bm{\theta}_\mathrm{F} \\ 
            \bm{0} & \bm{0} & \bm{0} 
        \end{array}
    \right] ,
    \quad
    \mathbf{V}_\times = \left[ 
        \begin{array}{ccc}
            \bm{0} & \bm{0} & c \bm{1} \\ 
            \bm{0} & \bm{0} & \bm{0} 
        \end{array}
    \right] , 
    \quad
    \bm{\alpha}_\times = \left[ 
        \begin{array}{c}
            \bm{\alpha}_\mathrm{F} \\ 
            \bm{0} 
        \end{array}
    \right] . 
\end{equation*}
Thus, one has 
\begin{equation}
    \label{eq: in the proof of ua of f-ftnet, f-ftnet equals fnn}
    \begin{aligned}
        f_{\times,\mathrm{F}}(\bm{x}) 
        =&~ \bm{\alpha}_\times^\top \mathrm{Re} \left[ \sigma_\times \left( \left( \mathbf{W}_{\mathrm{\times}} + \mathbf{V}_{\mathrm{\times}} \mathrm{i} \right) \kappa(\bm{x} , H_\times) \right) \right] \\ 
        =&~ \bm{\alpha}_\times^\top \mathrm{Re} \left[ \sigma_\times \left( \left[ \mathbf{W}_\mathrm{F} \bm{x} - \bm{\theta}_\mathrm{F} + c \bm{1} \mathrm{i} ; \bm{0} \right] \right) \right] \\ 
        =&~ \bm{\alpha}_\mathrm{F}^\top \mathrm{Re} \left[ \sigma_\times \left( \mathbf{W}_\mathrm{F} \bm{x} - \bm{\theta}_\mathrm{F} + c \bm{1} \mathrm{i} \right) \right] \\ 
        =&~ \bm{\alpha}_\mathrm{F}^\top \sigma \left( \mathbf{W}_\mathrm{F} \bm{x} - \bm{\theta}_\mathrm{F} \right) , 
    \end{aligned}
\end{equation}
where the first equality holds according to Eq.~\eqref{eq: mapping of f-ftnet}, the second and third equalities hold from the construction of the $\mathrm{F}\text{-}\mathrm{FTNet}$, and the fourth equality holds because of the definition of the function $\sigma$. From Eqs.~\eqref{eq: in the proof of ua of f-ftnet, ua of fnn} and~\eqref{eq: in the proof of ua of f-ftnet, f-ftnet equals fnn}, we obtain 
\begin{equation*} 
    \left\Vert f_{\times,\mathrm{F}}(\bm{x}) - g(\bm{x}) \right\Vert_{L^\infty(K)} 
    \leqslant \varepsilon , 
\end{equation*}
which completes the proof. $\hfill \square$ 

\subsection{Universal Approximation of \emph{R-FTNet}}
\label{subsec: universal approximation of r-ftnet}

We proceed to study the universal approximation for $\mathrm{R}\text{-}\mathrm{FTNet}$ as follows. 

\begin{theorem}
    \label{thm: ua of r-ftnet}
    Let $K \subset \mathbb{R}^{I}$ be a convex compact set, $f_\mathrm{D}$ is a $\mathrm{DODS}$ defined by Eq.~\eqref{eq: mapping of dods}, and $\sigma_\times$ is the activation function of $\mathrm{R}\text{-}\mathrm{FTNet}$ satisfying $\sigma_\times(0) = 0$. Suppose there exists a constant $c \in \mathbb{R}$, such that both $\sigma_1(x) = \mathrm{Re} [ \sigma_\times(x+c\mathrm{i}) ]$ and $\sigma_2(x) = \mathrm{Im} [ \sigma_\times(x+c\mathrm{i}) ]$ are continuous almost everywhere and not polynomials almost everywhere. Then for any $\varepsilon > 0$, there exists an $\mathrm{R}\text{-}\mathrm{FTNet}$ $f_{\times,\mathrm{R}}$, such that 
    \begin{equation*}
        \left\Vert f_{\times,\mathrm{R}} - f_\mathrm{D} \right\Vert_{L^\infty(K^T)} 
        \leqslant \varepsilon . 
    \end{equation*}
\end{theorem}

Theorem~\ref{thm: ua of r-ftnet} shows that $\mathrm{R}\text{-}\mathrm{FTNet}$ is a universal approximator. The requirement of convex domain $K$ is trivial since it is always possible to find a convex domain including a given compact domain. The conditions of the activation function are satisfied by many commonly used activation functions, such as $\mathrm{modReLU}$, $\mathrm{zReLU}$, and $\mathbb{C}\mathrm{ReLU}$. Existing studies investigate the universal approximation of $\mathrm{RNN}$, and the most general condition uses sigmoidal activation functions~\cite{schafer2006recurrent}. We extend the results to complex-weighted networks and generalize the requirement of activation functions.

\noindent \textbf{Proof of Theorem~\ref{thm: ua of r-ftnet}.} We start our proof with the universal approximation of an intermediate network, named additive $\mathrm{FTNet}$. One-hidden-layer additive $\mathrm{FTNet}$ with $H_+$ hidden neurons can be viewed as a mapping $f_{+,\mathrm{R}}$, defined by 
\begin{equation}
    \label{eq: mapping of addftnet, main body}
    \begin{aligned}
        & f_{+,\mathrm{R}} : \bm{x}_{1:T} \mapsto (y_{+,1},\dots,y_{+,T}) , \\ 
        & \bm{p}_t = \sigma_1 ( \mathbf{A} \bm{x}_t + \mathbf{B} \bm{q}_{t-1} - \bm{\zeta} ) , \\ 
        & \bm{q}_t = \sigma_2 ( \mathbf{A} \bm{x}_t + \mathbf{B} \bm{q}_{t-1} - \bm{\zeta} ) , \\ 
        & y_{+,t} = \bm{\alpha}_+^\top \bm{p}_t , \quad t \in [T] , 
    \end{aligned}
\end{equation}
where $ \mathbf{A} \in \mathbb{R}^{H_+ \times I} , \mathbf{B} \in \mathbb{R}^{H_+ \times H_+} , \bm{\alpha}_+ , \bm{\zeta} \in \mathbb{R}^{H_+} $ indicate weight parameters, and $\bm{p}_t , \bm{q}_t \in \mathbb{R}^{H_+} $ denote hidden states. We claim that there exists an additive $\mathrm{FTNet}$ $f_{+,\mathrm{R}}$, such that
\begin{equation}
    \label{eq: ua of addftnet, main body}
    \left\Vert f_{+,\mathrm{R}} - f_\mathrm{D} \right\Vert_{L^\infty(K^T)} 
    \leqslant \varepsilon . 
\end{equation}
This claim indicates the universal approximation of additive $\mathrm{FTNet}$. The proof of Eq.~\eqref{eq: ua of addftnet, main body} is similar to that of the universal approximation of $\mathrm{RNN}$ and is provided in Appendix. 

Based on Eq.~\eqref{eq: ua of addftnet, main body}, it suffices to prove that any additive $\mathrm{FTNet}$ using induced activation functions $\sigma_1$ and $\sigma_2$ is equivalent to an $\mathrm{R}\text{-}\mathrm{FTNet}$ using the activation function $\sigma$. 

Firstly, provided an additive $\mathrm{FTNet}$, the $\mathrm{R}\text{-}\mathrm{FTNet}$ with $H_\times = I + H_+ + 1$ hidden neurons is constructed as follows 
\begin{equation}
    \label{eq: in proof of ua of r-ftnet, construction of r-ftnet, main body}
    \mathbf{W}_\times = \left[ 
        \begin{array}{ccc}
            \mathbf{0}^{I \times I} & \mathbf{0} & \mathbf{0} \\ 
            \mathbf{0} & \mathbf{B} & - c \bm{1} \\ 
            \mathbf{0} & \mathbf{0} & \mathbf{0}^{1 \times 1} 
        \end{array}
    \right] , 
    \bm{r}_0 = \left[ 
        \begin{array}{c}
            \mathbf{0} \\ 
            \bm{q}_0 \\ 
            \mathbf{0} 
        \end{array}
    \right] ,
    \mathbf{V}_\times = \left[ 
        \begin{array}{ccc}
            \mathbf{0} & \mathbf{0} & \mathbf{0} \\ 
            \mathbf{A} & \mathbf{0} & - \bm{\zeta} \\ 
            \mathbf{0} & \mathbf{0} & \mathbf{0} 
        \end{array}
    \right] , 
    \bm{\alpha}_\times = \left[ 
        \begin{array}{c}
            \mathbf{0} \\ 
            \bm{\alpha}_+ \\ 
            \mathbf{0} 
        \end{array}
    \right] . 
\end{equation}

Secondly, we calculate the receptor, stimulus, and output of the above R-FTNet. We prove $ \bm{r}_t = [ \mathbf{0}^{I \times 1} ; \bm{q}_t ; \mathbf{0}^{1 \times 1} ] $ by mathematical induction as follows. 
\begin{enumerate}
    \item For $t=0$, the conclusion holds according to Eq.~\eqref{eq: in proof of ua of r-ftnet, construction of r-ftnet, main body}. 
    \item Suppose that the conclusion holds for $t = \tau$ with $\tau \leqslant T-1$. Thus, one has 
    \begin{equation*}
        \begin{aligned}
            \bm{r}_{\tau+1} 
            =&~ \mathrm{Im} \left[ \sigma_{\mathrm{\times}} \left( \mathbf{0} ; c \bm{1} + \left( \mathbf{A} \bm{x}_{\tau+1} + \mathbf{B} \bm{q}_\tau - \bm{\zeta} \right) \mathrm{i} ; \mathbf{0} \right) \right] \\ 
            =&~ \left[ \mathbf{0} ; \sigma_2 ( \mathbf{A} \bm{x}_{\tau+1} + \mathbf{B} \bm{q}_\tau - \bm{\zeta} ) ; \mathbf{0} \right] \\ 
            =&~ \left[ \mathbf{0} ; \bm{q}_{\tau+1} ; \mathbf{0} \right] , 
        \end{aligned}
    \end{equation*}
    where the first equality holds from Eqs.~\eqref{eq: mapping of r-ftnet},~\eqref{eq: in proof of ua of r-ftnet, construction of r-ftnet, main body} and the induction hypothesis, the second equality holds based on the definition of the activation function $\sigma_2$ in Theorem~\ref{thm: ua of r-ftnet}, and the third equality holds according to Eq.~\eqref{eq: mapping of addftnet, main body}. Thus, the conclusion holds for $t = \tau + 1$. 
\end{enumerate}
For any $t \in [T]$, the stimulus satisfies 
\begin{equation*}
    \begin{aligned}
        \bm{s}_t 
        =&~ \mathrm{Re} \left[ \sigma_{\mathrm{\times}} \left( \mathbf{0} ; c \bm{1} + \left( \mathbf{A} \bm{x}_t + \mathbf{B} \bm{q}_{t-1} - \bm{\zeta} \right) \mathrm{i} ; \mathbf{0} \right) \right] \\ 
        =&~ \left[ \mathbf{0} ; \sigma_2 ( c \bm{1} , \mathbf{A} \bm{x}_t + \mathbf{B} \bm{q}_{t-1} - \bm{\zeta} ) ; \mathbf{0} \right] \\ 
        =&~ \left[ \mathbf{0} ; \bm{p}_t ; \mathbf{0} \right] , 
    \end{aligned}
\end{equation*}
which leads to $y_{\times,t} = \bm{\alpha}_\times^\top \bm{s}_t = \bm{\alpha}_+^\top \bm{p}_t = y_{+,t}$. Therefore, the $\mathrm{R}\text{-}\mathrm{FTNet}$ defined by Eq.~\eqref{eq: in proof of ua of r-ftnet, construction of r-ftnet, main body} has the same output as the additive $\mathrm{FTNet}$ defined by Eq.~\eqref{eq: mapping of addftnet, main body}, i.e., 
\begin{equation}
    \label{eq: mulftnet is equivalent to addftnet, main body}
    f_{\times,\mathrm{R}}(\bm{x}_{1:T}) 
    = f_{+,\mathrm{R}}(\bm{x}_{1:T}) , 
    \quad 
    \forall ~ \bm{x}_{1:T} \in K^T . 
\end{equation}

Finally, from Eqs.~\eqref{eq: ua of addftnet, main body} and~\eqref{eq: mulftnet is equivalent to addftnet, main body}, one has 
\begin{equation*}
    \Vert f_{\times,\mathrm{R}}(\bm{x}_{1:T}) - f_\mathrm{D}(\bm{x}_{1:T}) \Vert_{L^\infty(K^T)}
    = \Vert f_{+,\mathrm{R}}(\bm{x}_{1:T}) - f_\mathrm{D}(\bm{x}_{1:T}) \Vert_{L^\infty(K^T)} 
    \leqslant \varepsilon , 
\end{equation*}
which completes the proof. $\hfill \square$

\section{Approximation Complexity}
\label{sec: approximation complexity}

We show the approximation advantage of $\mathrm{FTNet}$ over $\mathrm{FNN}$ and $\mathrm{RNN}$ in Subsection~\ref{subsec: advantage of mulftnet} and provide worst-case guarantees in Subsection~\ref{subsec: worst case guarantee of mulftnet}. Let us introduce the $(\varepsilon,\mathcal{D})$-approximation, which is used throughout this section. 

\begin{definition}
    \label{def: e,d approximation}
    Let $g$ be a function from $\mathbb{R}^I$ to $\mathbb{R}$, $\mathcal{F}$ is a class of functions from $\mathbb{R}^I$ to $\mathbb{R}$, and $\mathcal{D}$ is a distribution over $\mathbb{R}^I$. The function $g$ can be $(\varepsilon,\mathcal{D})$-approximated by function class $\mathcal{F}$ if there exists a function $f \in \mathcal{F}$, such that 
    \begin{equation*}
        \mathbb{E}_{\bm{x} \sim \mathcal{D}} \left[ \left( g(\bm{x}) - f(\bm{x}) \right)^2 \right] \leqslant \varepsilon . 
    \end{equation*}
\end{definition}

The $(\varepsilon,\mathcal{D})$-approximation means that the minimal expected squared difference between a function from the function class $\mathcal{F}$ and the target function $g$ is small. Let $\mathcal{F}$ be the function space of a neural network, and $g$ is the learning target. Then the $(\varepsilon,\mathcal{D})$-approximation indicates that it is possible to find a set of parameters for the neural network, such that the neural network suffers a negligible loss under the task of learning $g$. 

\subsection{Approximation-Complexity Advantage of \emph{FTNet}}
\label{subsec: advantage of mulftnet}

We now present two theorems showing the separation results between $\mathrm{FTNet}$ and $\mathrm{FNN}$/$\mathrm{RNN}$, respectively. 

\begin{theorem}
    \label{thm: advantage of ftnet over fnn}
    There exist constants $I_1 \in \mathbb{N}^+$, $\varepsilon_1 > 0$, and $c_1 > 0$, such that for any input dimension $I \geqslant I_1$, there exist a distribution $\mathcal{D}_1$ over $\mathbb{R}^I$ and a function $f_1 : \mathbb{R}^I \rightarrow \mathbb{R}$, s.t. 
    \begin{enumerate}
        \item For any $ \varepsilon > 0 $, the target function $f_1$ can be $(\varepsilon,\mathcal{D}_1)$-approximated by one-hidden-layer $\mathrm{F}\text{-}\mathrm{FTNet}$ with at most $ \max \{ 3 c_1^2 I^{15/2} / \varepsilon^2 , 27 I^2 \} $ parameters using the $\mathrm{zReLU}$ activation function. 
        \item The target function $f_1$ cannot be $(\varepsilon_1,\mathcal{D}_1)$-approximated by one-hidden-layer $\mathrm{FNN}$ with at most $\varepsilon_1 \mathrm{e}^{\varepsilon_1 I}$ parameters using the $\mathrm{ReLU}$ activation function. 
    \end{enumerate}
\end{theorem}

\begin{theorem}
    \label{thm: advantage of ftnet over rnn}
    There exist constants $I_2 \in \mathbb{N}^+$, $\varepsilon_2 > 0$, and $c_2 > 0$, such that for any input dimension $I \geqslant I_2$, there exist a distribution $\mathcal{D}_2$ over $\mathbb{R}^I$ and a $\mathrm{DODS}$ $f_\mathrm{D} : \mathbb{R}^{IT} \rightarrow \mathbb{R}^T$, s.t. 
    \begin{enumerate}
        \item For any $ \varepsilon > 0 $, the $\mathrm{DODS}$ $f_\mathrm{D}$ can be $(T \varepsilon,\mathcal{D}_2^T)$-approximated by one-hidden-layer $\mathrm{R}\text{-}\mathrm{FTNet}$ with at most $ 3 ( c_2 I^{15/4} / \varepsilon + 3I )^2 $ parameters using the $\mathrm{zReLU}$ activation function. 
        \item The $\mathrm{DODS}$ $f_\mathrm{D}$ cannot be $(T \varepsilon_2,\mathcal{D}_2^T)$-approximated by one-hidden-layer $\mathrm{RNN}$ with at most $ \varepsilon_0 \mathrm{e}^{\varepsilon_2 I} / 4 $ parameters using the $\mathrm{ReLU}$ activation function.
    \end{enumerate}
\end{theorem}

Theorems~\ref{thm: advantage of ftnet over fnn}~and~\ref{thm: advantage of ftnet over rnn} show the approximation-complexity advantage of $\mathrm{FTNet}$ over $\mathrm{FNN}$ and $\mathrm{RNN}$, respectively, i.e., there exists a target function such that $\mathrm{FTNet}$ can express it with polynomial parameters, but $\mathrm{FNN}$ or $\mathrm{RNN}$ cannot approximate it unless exponential parameters are used. Previous studies usually demonstrate separation results between deep networks and shallow networks~\cite{eldan2016power,safran2017depth,telgarsky2016benefits}. A recent study shows exponential separation between one-hidden-layer CRNet and one-hidden-layer $\mathrm{FNN}$~\cite{zhang2022towards}. Our results consider both feedforward and recurrent architectures and demonstrate the advantage of $\mathrm{FTNet}$ by showing that it is sufficient for one-hidden-layer $\mathrm{FTNet}$ to possess exponential separation over $\mathrm{FNN}$ and $\mathrm{RNN}$. We begin our proof by introducing the complex-reaction network ($\mathrm{CRNet}$) and an important lemma.

The complex-reaction network ($\mathrm{CRNet}$) is a recently proposed neural network with complex-valued operations~\cite{zhang2022towards}. The real-valued input vector $\bm{x} = ( x_1 ; x_2 ; \dots ; x_I ) \in \mathbb{R}^I$ is folded by a transformation mapping $\tau : \mathbb{R}^I \rightarrow \mathbb{C}^{I/2}$ to form a complex-valued vector, i.e., 
\begin{equation*}
    \tau : \bm{x} \mapsto \left( x_1 ; x_2 ; \dots ; x_{I/2} \right) + \left( x_{I/2+1} ; x_{I/2+2} ; \dots ; x_{I/2} \right) \mathrm{i} , 
\end{equation*}
where the input dimension $I$ is assumed to be an even number without loss of generality. Recalling the formulation of $\mathrm{CRNet}$~\cite{zhang2022towards}, one-hidden-layer $\mathrm{CRNet}$ with $H_\mathrm{C}$ hidden neurons is a mapping $f_\mathrm{C} : \mathbb{R}^I \rightarrow \mathbb{R}$ of the following form 
\begin{equation}
    \label{eq: definition of crnet}
    f_{\mathrm{C}} : \bm{x} \mapsto \mathrm{Re} \left[ \bm{\alpha}_{\mathrm{C}}^\top \sigma_{\mathrm{C}} \left( \mathbf{W}_{\mathrm{C}} \tau(\bm{x}) + \bm{b}_{\mathrm{C}} \right) \right] , 
\end{equation}
where $\mathbf{W}_{\mathrm{C}} \in \mathbb{C}^{ H_{\mathrm{C}} \times d }$, $\bm{b}_{\mathrm{C}} \in \mathbb{C}^{H_{\mathrm{C}}}$, $\bm{\alpha}_{\mathrm{C}} \in \mathbb{C}^{H_{\mathrm{C}}}$ indicate weight parameters, and $\sigma_{\mathrm{C}} : \mathbb{C} \rightarrow \mathbb{C}$ is a complex-valued activation function applied componentwise. 

\begin{lemma}{\cite[Theorem 1]{eldan2016power}, \cite[Theorem 2]{zhang2022towards}}
    \label{lem: crnet vs fnn}
    There exist constants $I_0 \in \mathbb{N}^+$, $\varepsilon_0 > 0$, and $c_0 > 0$, such that for any input dimension $I \geqslant I_0$, there exist a distribution $\mathcal{D}_0$ over $\mathbb{R}^I$ and a function $f_0 : \mathbb{R}^I \rightarrow \mathbb{R}$, such that 
    \begin{enumerate}
        \item For any $\varepsilon > 0$, $f_0$ can be $(\varepsilon,\mathcal{D}_0)$-approximated by one-hidden-layer $\mathrm{CRNet}$ with at most $ c_0 I^{19/4} / \varepsilon $ parameters using the $\mathrm{zReLU}$ activation function. 
        \item The function $f_0$ cannot be $(\varepsilon_0,\mathcal{D}_0)$-approximated by one-hidden-layer $\mathrm{FNN}$ with at most $\varepsilon_0 \mathrm{e}^{\varepsilon_0 I}$ parameters using the $\mathrm{ReLU}$ activation function. 
    \end{enumerate}
\end{lemma}

Lemma~\ref{lem: crnet vs fnn} indicates the approximation-complexity advantage of $\mathrm{CRNet}$ over $\mathrm{FNN}$, i.e., there exists a target function such that $\mathrm{CRNet}$ can express it with polynomial parameters, but $\mathrm{FNN}$ cannot express it unless exponential parameters are used. 

\noindent \textbf{Proof of Theorem~\ref{thm: advantage of ftnet over fnn}.} Let $I_1 = I_0$, $\varepsilon_1 = \varepsilon_0$, and $c_1 = c_0$, where $I_0$, $\varepsilon_0$, and $c_0$ are defined in Lemma~\ref{lem: crnet vs fnn}. For any $I \geqslant I_1$, let $\mathcal{D}_1 = \mathcal{D}$ and $f_1 = f_0$. Without loss of generality, let the input dimension $I$ be an even number. 

Firstly, we prove that $\mathrm{F}\text{-}\mathrm{FTNet}$ can approximate the target function $f_1$ using polynomial parameters. Recalling the definition of $\mathrm{CRNet}$ in Eq.~\eqref{eq: definition of crnet}, we define 
\begin{equation}
  \label{eq: real valued weight for crnet}
  \begin{aligned}
    & \mathbf{W}_{\mathrm{C}} = \mathbf{W}_{\mathrm{C},R} + \mathbf{W}_{\mathrm{C},I} \mathrm{i} , \\
    & \bm{b}_{\mathrm{C}} = \bm{b}_{\mathrm{C},R} + \bm{b}_{\mathrm{C},I} \mathrm{i} , \\ 
    & \bm{\alpha}_{\mathrm{C}} = \bm{\alpha}_{\mathrm{C},R} + \bm{\alpha}_{\mathrm{C},I} \mathrm{i} , 
  \end{aligned}
\end{equation}
where $\mathbf{W}_{\mathrm{C},R} , \mathbf{W}_{\mathrm{C},I} \in \mathbb{R}^{H_{\mathrm{C}} \times I/2}$, $\bm{b}_{\mathrm{C},R} , \bm{b}_{\mathrm{C},I} \in \mathbb{R}^{H_{\mathrm{C}}}$, and $\bm{\alpha}_{\mathrm{C},R} , \bm{\alpha}_{\mathrm{C},I} \in \mathbb{R}^{H_{\mathrm{C}}}$ are real-valued parameters. The rest of the proof is divided into several steps. \\
\textbf{Step 1.} We construct an F-FTNet with the same output as a given CRNet. The $\mathrm{F}\text{-}\mathrm{FTNet}$ with $H_\times = \max \{ 2H_{\mathrm{C}} , I+1 \}$ hidden neurons is constructed as follows 
\begin{equation*}
    \begin{aligned}
        & \mathbf{W}_{\times} = \left[ 
            \begin{array}{cccc}
                \mathbf{W}_{\mathrm{C},R} & - \mathbf{W}_{\mathrm{C},I} & \mathbf{0} & \bm{b}_{\mathrm{C},R} \\ 
                \mathbf{W}_{\mathrm{C},I} & \mathbf{W}_{\mathrm{C},R} & \mathbf{0} & \bm{b}_{\mathrm{C},I} \\ 
                \mathbf{0} & \mathbf{0} & \mathbf{0} & \mathbf{0} 
            \end{array}
        \right] ,
        \quad
        \mathbf{V}_{\times} = \left[ 
            \begin{array}{cccc}
                \mathbf{W}_{\mathrm{C},I} & \mathbf{W}_{\mathrm{C},R} & \mathbf{0} & \bm{b}_{\mathrm{C},I} \\ 
                \mathbf{W}_{\mathrm{C},R} & - \mathbf{W}_{\mathrm{C},I} & \mathbf{0} & \bm{b}_{\mathrm{C},R} \\ 
                \mathbf{0} & \mathbf{0} & \mathbf{0} & \mathbf{0} 
            \end{array}
        \right] , \\
        & \bm{r}_0 = \mathbf{0} , 
        \quad 
        \bm{\alpha}_{\times} = \left[ \bm{\alpha}_{\mathrm{C},R} ; - \bm{\alpha}_{\mathrm{C},I} ; \mathbf{0} \right] ,
    \end{aligned}
\end{equation*}
and $\sigma_{\times}$ is the $\mathrm{zReLU}$ activation function. The output of the constructed $\mathrm{F}\text{-}\mathrm{FTNet}$ above satisfies 
\begin{equation}
    \label{eq: expression of f_times}
    \begin{aligned}
        f_{\times,\mathrm{F}}(\bm{x})
        &= \bm{\alpha}_{\times}^\top \mathrm{Re} \left[ \sigma_{\mathrm{\times}} \left( \left( \mathbf{W}_{\mathrm{\times}} + \mathbf{V}_{\mathrm{\times}} \mathrm{i} \right) \left( \kappa(\bm{x} , H_\times) + \bm{r}_0 \mathrm{i} \right) \right) \right] \\ 
        &= \bm{\alpha}_{\times}^\top \mathrm{Re} \left[ \sigma_{\mathrm{C}} \left( \left[ \mathbf{W}_{\mathrm{C}} \tau(\bm{x}) + \bm{b}_{\mathrm{C}} ; \overline{ \mathbf{W}_{\mathrm{C}} \tau(\bm{x}) + \bm{b}_{\mathrm{C}} } \mathrm{i} ; \mathbf{0} \right] \right) \right] \\ 
        &= \bm{\alpha}_{\mathrm{C},R}^\top \mathrm{Re} \left[ \sigma_{\mathrm{C}} \left( \mathbf{W}_{\mathrm{C}} \tau(\bm{x}) + \bm{b}_{\mathrm{C}} \right) \right] - \bm{\alpha}_{\mathrm{C},I}^\top \mathrm{Re} \left[ \sigma_{\mathrm{C}} \left( \overline{ \mathbf{W}_{\mathrm{C}} \tau(\bm{x}) + \bm{b}_{\mathrm{C}} } \mathrm{i} \right) \right] , 
    \end{aligned}
\end{equation}
It is observed that 
\begin{equation*}
    \mathrm{Re} \left[ \sigma_{\mathrm{C}}(x+y\mathrm{i}) \right] 
    = \mathrm{Im} \left[ \sigma_{\mathrm{C}}(\overline{x+y\mathrm{i}} ~ \mathrm{i}) \right] 
    = \left\{ 
        \begin{array}{ll}
            x, & \mathrm{if} ~ xy \geqslant 0 , \\ 
            0, & \mathrm{otherwise} . 
        \end{array}
    \right. 
\end{equation*}
Thus, one has 
\begin{equation}
    \label{eq: f_times equals f_C}
    \begin{aligned}
        f_{\times,\mathrm{F}}(\bm{x}) 
        =&~ \mathrm{Re} \left[ \left( \bm{\alpha}_{\mathrm{C},R} + \bm{\alpha}_{\mathrm{C},I} \mathrm{i} \right)^\top \sigma_{\mathrm{C}} \left( \mathbf{W}_{\mathrm{C}} \tau(\bm{x}) + \bm{b}_{\mathrm{C}} \right) \right] \\ 
        =&~ \mathrm{Re} \left[ \bm{\alpha}_{\mathrm{C}}^\top \sigma_{\mathrm{C}} \left( \mathbf{W}_{\mathrm{C}} \tau(\bm{x}) + \bm{b}_{\mathrm{C}} \right) \right] \\ 
        =&~ f_{\mathrm{C}} (\bm{x}) , 
    \end{aligned}
\end{equation}
which indicates that any $\mathrm{CRNet}$ with hidden size $H_\mathrm{C}$ can be expressed by an $\mathrm{F}\text{-}\mathrm{FTNet}$ with hidden size $\max \{ 2H_{\mathrm{C}} , I+1 \}$. \\ 
\textbf{Step 2.} We bound the number of required parameters in the constructed $\mathrm{F}\text{-}\mathrm{FTNet}$. From Lemma~\ref{lem: crnet vs fnn}, for any $ \varepsilon > 0 $, there exists $\mathrm{CRNet}$ $ f_{\mathrm{C}} $ with at most $ ( c_1 I^{19/4} ) / \varepsilon $ parameters using the $\mathrm{zReLU}$ activation function, such that 
\begin{equation}
    \label{eq: crnet approximate f1}
    \mathbb{E}_{\bm{x} \sim \mathcal{D}} \left[ \left( f_{\mathrm{C}}(\bm{x}) - f_1(\bm{x}) \right)^2 \right] 
    \leqslant \varepsilon . 
\end{equation}
For $\mathrm{CRNet}$ with $H_{\mathrm{C}}$ hidden neurons, it has $2 H_{\mathrm{C}} (I+2)$ parameters. Thus, the hidden size of $\mathrm{CRNet}$ satisfies 
\begin{equation*}
    H_{\mathrm{C}} \leqslant \frac{ c_1 I^{19/4} }{ 2 (I+2) \varepsilon } \leqslant \frac{ c_1 I^{15/4} }{ 2 \varepsilon } . 
\end{equation*}
According to Step 2, there exists an $\mathrm{F}\text{-}\mathrm{FTNet}$, with no more than $\max \{ 2H_\mathrm{C},I+1 \}$ hidden neurons, satisfying $f_{\times,\mathrm{F}}(\bm{x}) = f_\mathrm{C}(\bm{x})$. This property, together with Eq.~\eqref{eq: crnet approximate f1}, indicates that 
\begin{equation*}
    \mathbb{E}_{\bm{x} \sim \mathcal{D}} \left[ \left( f_{\times,\mathrm{F}}(\bm{x}) - f_1(\bm{x}) \right)^2 \right] 
    \leqslant \varepsilon , 
\end{equation*}
and that the number of hidden neurons in the constructed $\mathrm{F}\text{-}\mathrm{FTNet}$ $f_{\times,\mathrm{F}}$ is no more than 
\begin{equation*}
    H_{\times} 
    \leqslant \max \left\{ 2 H_{\mathrm{C}} , I+1 \right\} 
    \leqslant \max \left\{ c_1 I^{15/4} / \varepsilon , I+1 \right\} . 
\end{equation*}
For $\mathrm{F}\text{-}\mathrm{FTNet}$ with $H_{\times}$ hidden neurons, it has $2 H_{\times}^2 + H_{\times}$ parameters. Thus, the number of parameters in the constructed $\mathrm{F}\text{-}\mathrm{FTNet}$ $f_{\times,\mathrm{F}}$ is no more than 
\begin{equation*}
    2 H_{\times}^2 + H_{\times} 
    \leqslant 3 H_{\times}^2 
    \leqslant \max \left\{ 3 c_1^2 I^{15/2} / \varepsilon^2 , 27 I^2 \right\} . 
\end{equation*}

Secondly, Lemma~\ref{lem: crnet vs fnn} indicates that $\mathrm{FNN}$ needs at least exponential parameters to approximate the target function $f_1$. 

Combining the conclusions above completes the proof. $\hfill \square$ 

\noindent \textbf{Proof of Theorem~\ref{thm: advantage of ftnet over rnn}.} Let $I_2 = I_0$, $\varepsilon_2 = \varepsilon_0$, and $c_2 = c_0$, where $I_0$, $\varepsilon_0$, and $c_0$ are defined in Lemma~\ref{lem: crnet vs fnn}. For any $I \geqslant I_2$, let $\mathcal{D}_2 = \mathcal{D}$. Without loss of generality, let the input dimension $I$ be an even number. The $\mathrm{DODS}$ is constructed as follows. For any input $\bm{x} \in \mathbb{R}^I$ and hidden state $\bm{h} \in \mathbb{R}^{H_\mathrm{D}}$, let $\varphi(\bm{x},\bm{h}) = \bm{x}$ and $\psi(\bm{h}) = f_0(\bm{h})$, where $f_0$ is the same function as that in Lemma~\ref{lem: crnet vs fnn}. Thus, the output at time $t$ is 
\begin{equation}
    \label{eq: output of dods, in ftnet vs rnn}
    y_t 
    = \psi(\bm{h}_t) 
    = \psi(\varphi(\bm{x}_t,\bm{h}_{t-1})) 
    = f_0(\bm{x}_t) , 
\end{equation}
which holds according to Eq.~\eqref{eq: mapping of dods}. 

Firstly, we prove that $\mathrm{R}\text{-}\mathrm{FTNet}$ can express $f_\mathrm{D}$ using polynomial parameters. The proof is divided into several steps. \\ 
\textbf{Step 1.} We construct an $\mathrm{R}\text{-}\mathrm{FTNet}$ with the same output as a given CRNet. From the proof of Theorem~\ref{thm: advantage of ftnet over fnn}, for any $ \varepsilon > 0 $, there exists a $\mathrm{CRNet}$ $ f_{\mathrm{C}} $ with at most $ H_{\mathrm{C}} = ( c_1 I^{15/4} ) / ( 2 \varepsilon ) $ hidden neurons using the $\mathrm{zReLU}$ activation function, such that 
\begin{equation}
    \label{eq: crnet approximate f}
    \mathbb{E}_{\bm{x} \sim \mathcal{D}} \left[ \left( f_{\mathrm{C}}(\bm{x}) - f_0(\bm{x}) \right)^2 \right] 
    \leqslant \varepsilon . 
\end{equation}
Let $\mathbf{W}_{\mathrm{C},R}$, $\mathbf{W}_{\mathrm{C},I}$, $\bm{b}_{\mathrm{C},R}$, $\bm{b}_{\mathrm{C},I} $, $\bm{\alpha}_{\mathrm{C},R}$, and $\bm{\alpha}_{\mathrm{C},I}$ be the real-valued weight matrices of the above $\mathrm{CRNet}$, which are defined in the same way as those in Eq.~\eqref{eq: real valued weight for crnet}. Define the $\mathrm{R}\text{-}\mathrm{FTNet}$ $f_{\times,\mathrm{R}}$ with $H_\times = 2H_{\mathrm{C}} + I + 1$ hidden neurons as follows 
\begin{equation*}
    \begin{aligned}
        & \mathbf{W}_{\times} = \left[ 
            \begin{array}{cccc}
                \mathbf{0}^{I \times I/2} & \mathbf{0} & \mathbf{0} & \mathbf{0} \\ 
                \mathbf{W}_{\mathrm{C},R} & - \mathbf{W}_{\mathrm{C},I} & \mathbf{0} & \bm{b}_{\mathrm{C},R} \\ 
                \mathbf{W}_{\mathrm{C},I} & \mathbf{W}_{\mathrm{C},R} & \mathbf{0} & \bm{b}_{\mathrm{C},I} \\ 
                \mathbf{0} & \mathbf{0} & \mathbf{0} & \mathbf{0} 
            \end{array}
        \right] ,
        \quad
        \mathbf{V}_{\times} = \left[ 
            \begin{array}{cccc}
                \mathbf{0}^{I \times I/2} & \mathbf{0} & \mathbf{0} & \mathbf{0} \\ 
                \mathbf{W}_{\mathrm{C},I} & \mathbf{W}_{\mathrm{C},R} & \mathbf{0} & \bm{b}_{\mathrm{C},I} \\ 
                \mathbf{W}_{\mathrm{C},R} & - \mathbf{W}_{\mathrm{C},I} & \mathbf{0} & \bm{b}_{\mathrm{C},R} \\ 
                \mathbf{0} & \mathbf{0} & \mathbf{0} & \mathbf{0} 
            \end{array}
        \right] , \\ 
        & \bm{r}_0 = \mathbf{0} , 
        \quad 
        \bm{\alpha}_{\times} = \left[ \mathbf{0} ; \bm{\alpha}_{\mathrm{C},R} ; - \bm{\alpha}_{\mathrm{C},I} ; \mathbf{0} \right] ,
    \end{aligned}
\end{equation*}
and $\sigma_{\times}$ is the $\mathrm{zReLU}$ activation function. We then prove that the output of the above $\mathrm{R}\text{-}\mathrm{FTNet}$ is the same as that of the $\mathrm{CRNet}$ in Eq.~\eqref{eq: crnet approximate f}. Since the first $I$ rows and the last row in $\mathbf{W}_{\times}$ and $\mathbf{V}_{\times}$ are all $0$, one has $\bm{r}_t = [ \mathbf{0}_{I \times 1} ; \tilde{\bm{r}}_t ; \mathbf{0}_{1 \times 1} ]$ for any $t \in [T]$, where $\tilde{\bm{r}}_t \in \mathbb{R}^{2 H_{\mathrm{C}}}$ is an arbitrary vector. Then the output of the $\mathrm{R}\text{-}\mathrm{FTNet}$ at time $t$ is 
\begin{equation*}
    \begin{aligned}
        y_{\times,t} 
        =& \bm{\alpha}_{\times}^\top \mathrm{Re} \left[ \sigma_{\mathrm{\times}} \left( \left( \mathbf{W}_{\mathrm{\times}} + \mathbf{V}_{\mathrm{\times}} \mathrm{i} \right) \left( \kappa(\bm{x}_t , H_\times) + \bm{r}_{t-1} \mathrm{i} \right) \right) \right] \\ 
        =&~ \bm{\alpha}_{\mathrm{C},R}^\top \mathrm{Re} \left[ \sigma_{\mathrm{C}} \left( \mathbf{W}_{\mathrm{C}} \tau(\bm{x}_t) + \bm{b}_{\mathrm{C}} \right) \right] - \bm{\alpha}_{\mathrm{C},I}^\top \mathrm{Re} \left[ \sigma_{\mathrm{C}} \left( \overline{ \mathbf{W}_{\mathrm{C}} \tau(\bm{x}_t) + \bm{b}_{\mathrm{C}} } \mathrm{i} \right) \right] . 
    \end{aligned}
\end{equation*}
The right-hand side of the above equation is the same as that of Eq.~\eqref{eq: expression of f_times}, except substituting $\bm{x}$ with $\bm{x}_t$. By similar derivation used in Eq.~\eqref{eq: f_times equals f_C}, one has 
\begin{equation}
    \label{eq: mulftnet equals crnet, in ftnet vs rnn}
    y_{\times,t} = f_{\mathrm{C}}(\bm{x}_t) , 
    \quad 
    \forall ~ t \in [T] . 
\end{equation}
\textbf{Step 2.} We prove that the $\mathrm{R}\text{-}\mathrm{FTNet}$ constructed in Step 1 can approximate $\mathrm{DODS}$ $f_\mathrm{D}$ with a small expected squared loss and then bound the number of required parameters. The expected squared loss of the above $\mathrm{R}\text{-}\mathrm{FTNet}$ is 
\begin{equation*}
    \begin{aligned}
        \mathbb{E}_{\bm{x}_{1:T} \sim \mathcal{D}_2^T} \left[  \left\Vert f_{\times,\mathrm{R}}(\bm{x}_{1:T}) - f_\mathcal{D}(\bm{x}_{1:T}) \right\Vert^2 \right]
        &= \mathbb{E}_{\bm{x}_{1:T} \sim \mathcal{D}_2^T} \left[ \sum_{t=1}^T \left( y_{\times,t} - y_t \right)^2 \right] \\ 
        &= \mathbb{E}_{\bm{x}_{1:T} \sim \mathcal{D}_2^T} \left[ \sum_{t=1}^T \left( f_{\mathrm{C}}(\bm{x}_t) - f_0(\bm{x}_t) \right)^2 \right] \\ 
        &\leqslant T \varepsilon , 
    \end{aligned}
\end{equation*}
where the second equality holds from Eqs.~\eqref{eq: output of dods, in ftnet vs rnn},~\eqref{eq: mulftnet equals crnet, in ftnet vs rnn}, and the first inequality holds based on Eq.~\eqref{eq: crnet approximate f}. We then calculate the number of parameters in the above $\mathrm{R}\text{-}\mathrm{FTNet}$. Since $\mathrm{FTNet}$ with hidden size $H_{\times}$ has $2 H_{\times}^2 + H_{\times}$ parameters, the number of parameters in the constructed $\mathrm{R}\text{-}\mathrm{FTNet}$ is no more than 
\begin{equation*}
    2 H_{\times}^2 + H_{\times} 
    \leqslant 3 H_{\times}^2 
    \leqslant 3 \left( c_1 I^{15/4} / \varepsilon + 3I \right)^2 . 
\end{equation*}

Secondly, we prove that $\mathrm{RNN}$ needs at least exponential parameters to approximate the target $\mathrm{DODS}$ $f_\mathrm{D}$. The proof is divided into several steps. \\ 
\textbf{Step 1.} We prove that if the total loss suffered by $\mathrm{RNN}$ is large, there exists a time point $t \in [T]$, such that $\mathrm{RNN}$ suffers a large loss at time $t$. For the unity of notations, we rewrite the one-hidden-layer $\mathrm{RNN}$ $f_\mathrm{R}$ with $H_\mathrm{R}$ hidden neurons as follows 
\begin{equation}
    \label{eq: definition of rnn}
    \begin{aligned}
        & f_\mathrm{R} : \bm{x}_{1:T} \mapsto (y_{\mathrm{R},1},\dots,y_{\mathrm{R},T}) , \\ 
        & \bm{m}_t = \sigma_\mathrm{R} \left( \mathbf{W}_\mathrm{R} \bm{x}_t + \mathbf{V}_\mathrm{R} \bm{m}_{t-1} - \bm{\zeta}_\mathrm{R} \right) , \\ 
        & y_{\mathrm{R},t} = \bm{\alpha}_{\mathrm{R}}^\top \bm{m}_t , \quad \mathrm{for} \quad t \in [T] , 
    \end{aligned}
\end{equation}
where $\bm{m}_t \in \mathbb{R}^{H_\mathrm{R}}$ and $y_{\mathrm{R},t} \in \mathbb{R}$ represent the memory and output at time $t$, respectively, $\mathbf{W}_\mathrm{R}$, $\mathbf{V}_\mathrm{R}$, $\bm{\zeta}_\mathrm{R}$, $\bm{\alpha}_\mathrm{R}$ denote weight parameters, and $\sigma_\mathrm{R}$ is the $\mathrm{ReLU}$ activation function applied componentwise. If the $\mathrm{DODS}$ can be $(T \varepsilon_0,\mathcal{D}_2^T)$-approximated by $\mathrm{RNN}$, the following holds from Definition~\ref{def: e,d approximation},
\begin{equation*}
T \varepsilon_0 
\geqslant \mathbb{E}_{\bm{x}_{1:T} \sim \mathcal{D}_2^T} \left[ \left\Vert f_\mathrm{R}(\bm{x}_{1:T}) - f_\mathrm{D}(\bm{x}_{1:T}) \right\Vert^2 \right]
= \mathbb{E}_{\bm{x}_{1:T} \sim \mathcal{D}_2^T} \left[ \sum\nolimits_{t=1}^T \left( y_{\mathrm{R},t} - y_t \right)^2 \right] . 
\end{equation*}
Since for any time $t \in [T]$, the squared term $( y_{\mathrm{R},t} - y_t )^2$ is always non-negative, there exists time $t_0 \in [T]$, such that $\mathbb{E}_{\bm{x}_{t_0} \sim \mathcal{D}_2} [ ( y_{\mathrm{R},t_0} - y_{t_0} )^2 ] \leqslant \varepsilon_0$. According to the definitions of $y_{\mathrm{R},t_0}$ and $y_{t_0}$ in Eqs.~\eqref{eq: definition of rnn}~and~\eqref{eq: output of dods, in ftnet vs rnn}, one has 
\begin{equation}
    \label{eq: error of rnn at a time point is small}
    \mathbb{E}_{\bm{x} \sim \mathcal{D}_2} \bigg[ \big( \bm{\alpha}_{\mathrm{R}}^\top \sigma_{\mathrm{R}} ( \mathbf{W}_{\mathrm{R}} \bm{x} + \mathbf{V}_{\mathrm{R}} \bm{m}_{t_0-1} + \bm{\theta}_{\mathrm{R}} ) - f_0(\bm{x}) \big)^2 \bigg] \leqslant \varepsilon_0 . 
\end{equation}
\textbf{Step 2.} We use Lemma~\ref{lem: crnet vs fnn} to give a lower bound on the number of parameters of $\mathrm{RNN}$ with small loss. Before the proof, we rewrite the $\mathrm{FNN}$ in the mapping form for the unity of notation. One-hidden-layer $\mathrm{FNN}$ with $H_\mathrm{F}$ hidden neurons can be viewed as a mapping $f_\mathrm{F}$, defined by 
\begin{equation}
    \label{eq: definition of fnn}
    f_\mathrm{F} : \bm{x} \mapsto \bm{\alpha}_{\mathrm{R}}^\top \sigma_\mathrm{F} \left( \mathbf{W}_\mathrm{F} \bm{x} - \bm{\zeta}_\mathrm{F} \right) , 
\end{equation}
where $\bm{x} \in \mathbb{R}^I$ represents input at time $t$, $\mathbf{W}_\mathrm{F} \in \mathbb{R}^{H_\mathrm{F} \times I}$, $\bm{\zeta}_\mathrm{F}$, $\bm{\alpha}_\mathrm{F} \in \mathbb{R}^{H_\mathrm{F}}$ denote weight parameters, and $\sigma_\mathrm{F}$ is the $\mathrm{ReLU}$ activation function applied componentwise. We now construct an $\mathrm{FNN}$ equivalent to the $\mathrm{RNN}$ at time $t_0$ as follows. Let $\bm{\alpha}_{\mathrm{F}} = \bm{\alpha}_{\mathrm{R}}$, $\mathbf{W}_{\mathrm{F}} = \mathbf{W}_{\mathrm{R}}$, and $\bm{b}_{\mathrm{F}} = \mathbf{V}_{\mathrm{R}} \bm{m}_{t_0-1} + \bm{\theta}_{\mathrm{R}}$. From Eq.~\eqref{eq: error of rnn at a time point is small}, one has $\mathbb{E}_{\bm{x} \sim \mathcal{D}_2} [ ( f_{\mathrm{F}}(\bm{x}) - f_0(\bm{x}) )^2 ] \leqslant \varepsilon_0$. According to Lemma~\ref{lem: crnet vs fnn}, the number of parameters of $f_{\mathrm{F}}$ is at least $\varepsilon_0 \mathrm{e}^{\varepsilon_0 I}$. For $\mathrm{FNN}$ with $H_{\mathrm{F}}$ hidden neurons, it has $2 H_{\mathrm{F}} (I+1)$ parameters. Thus, the hidden size of the $\mathrm{FNN}$ satisfies 
\begin{equation*}
    H_{\mathrm{F}} \geqslant \frac{ \varepsilon_0 \mathrm{e}^{\varepsilon_0 I} }{ 2 (I+1) } \geqslant \frac{ \varepsilon_0 \mathrm{e}^{\varepsilon_0 I} }{ 4I } . 
\end{equation*}
For $\mathrm{RNN}$ with $H_{\mathrm{R}}$ hidden neurons, it has $H_{\mathrm{R}} (I + H_{\mathrm{R}} + 2)$ parameters. Since the above $\mathrm{FNN}$ has the same hidden size as the $\mathrm{RNN}$, i.e., $H_{\mathrm{R}} = H_{\mathrm{F}}$, one knows that the number of parameters of $\mathrm{RNN}$ satisfies 
\begin{equation*}
    H_{\mathrm{R}} (I + H_{\mathrm{R}} + 2) 
    > H_{\mathrm{R}} I 
    = H_{\mathrm{F}} I 
    \geqslant \frac{ \varepsilon_0 \mathrm{e}^{\varepsilon_0 I} }{ 4 } . 
\end{equation*}

Combining the conclusions above completes the proof. $\hfill \square$ 

\subsection{Worst-Case Guarantee of \emph{FTNet}} 
\label{subsec: worst case guarantee of mulftnet}

We proceed to provide the worst-case guarantees of approximation complexity for $\mathrm{F}\text{-}\mathrm{FTNet}$ and $\mathrm{R}\text{-}\mathrm{FTNet}$. 

\begin{theorem}
    \label{thm: worst case guarantee of ftnet over fnn}
    Let $f$ be a function from $\mathbb{R}^I$ to $\mathbb{R}$, and $\mathcal{D}$ is a distribution over $\mathbb{R}^I$. For any $\varepsilon > 0$, if $f$ can be $(\varepsilon,\mathcal{D})$-approximated by one-hidden-layer $\mathrm{FNN}$ with hidden size $H_{\mathrm{F}}$ using the $\mathrm{ReLU}$ activation function, then $f$ can be $(\varepsilon,\mathcal{D})$-approximated by one-hidden-layer $\mathrm{F}\text{-}\mathrm{FTNet}_{\times}$ with hidden size $\max \{ H_{\mathrm{F}} , I+1 \}$ using the $\mathrm{zReLU}$ activation function. 
\end{theorem}

\begin{theorem}
    \label{thm: worst case guarantee of ftnet over rnn}
    Let $f_\mathrm{D}$ be a $\mathrm{DODS}$ from $\mathbb{R}^{IT}$ to $\mathbb{R}^T$, and $\mathcal{D}$ is a distribution over $\mathbb{R}^{TI}$. For any $\varepsilon > 0$, if $f_\mathrm{D}$ can be $(\varepsilon,\mathcal{D})$-approximated by one-hidden-layer $\mathrm{RNN}$ with hidden size $H_{\mathrm{R}}$ using the $\mathrm{ReLU}$ activation function, then $f_\mathrm{D}$ can be $(\varepsilon,\mathcal{D})$-approximated by one-hidden-layer $\mathrm{R}\text{-}\mathrm{FTNet}_{\times}$ with $2 H_{\mathrm{R}} + I + 1$ hidden neurons using the $\mathrm{zReLU}$ activation function. 
\end{theorem}

Theorems~\ref{thm: worst case guarantee of ftnet over fnn}~and~\ref{thm: worst case guarantee of ftnet over rnn} provide the worst-case guarantees for $\mathrm{FTNet}$, saying that the disadvantages of $\mathrm{FTNet}$ over $\mathrm{FNN}$ and $\mathrm{RNN}$ are no more than constants. Previous studies only provide separation advantages of model A over model B when expressing particular functions~\cite{eldan2016power,safran2017depth,telgarsky2016benefits,lu2017expressive,zhang2022towards}, without considering the opposite problem, i.e., whether model B possesses separation advantages over model A when approximating other functions. To our knowledge, our work is the first one to realize the opposite problem and provide a negative answer.

\noindent \textbf{Proof of Theorem~\ref{thm: worst case guarantee of ftnet over fnn}.} Since $f$ can be $(\varepsilon,\mathcal{D})$-approximated by $\mathrm{FNN}$, there exists an $\mathrm{FNN}$ defined by Eq.~\eqref{eq: definition of fnn}, such that 
\begin{equation}
    \label{eq: fnn approximate f, in guarantee of fnn}
    \mathbb{E}_{\bm{x} \sim \mathcal{D}} \left[ ( f(\bm{x}) - f_\mathrm{F}(\bm{x}) )^2 \right] 
    \leqslant \varepsilon . 
\end{equation}

Firstly, an $\mathrm{F}\text{-}\mathrm{FTNet}_{\times}$ $f_{\times,\mathrm{F}}$ with $H_{\times} = \max \{ H_{\mathrm{F}} , I+1 \}$ hidden neurons is constructed as follows 
\begin{equation*}
    \mathbf{W}_{\times} = \left[ 
        \begin{array}{ccc}
            \mathbf{W}_{\mathrm{F}} & \mathbf{0} & \bm{b}_{\mathrm{F}} \\ 
            \mathbf{0} & \mathbf{0} & \mathbf{0} 
        \end{array}
    \right] ,
    \quad
    \mathbf{V}_{\times} = \left[ 
        \begin{array}{ccc}
            \mathbf{0}^{} & \mathbf{0} & \bm{1} \\ 
            \mathbf{0} & \mathbf{0} & \mathbf{0} 
        \end{array}
    \right] ,
    \quad
    \bm{r}_0 = \mathbf{0} ,
    \quad
    \bm{\alpha}_{\times} = \left[
        \begin{array}{c}
            \bm{\alpha}_{\mathrm{F}} \\
            \mathbf{0}
        \end{array}
    \right] ,
\end{equation*}
and $\sigma_{\times}$ is the $\mathrm{zReLU}$ activation function.

Secondly, we prove that the output of the above $\mathrm{F}\text{-}\mathrm{FTNet}_{\times}$ is the same as that of the $\mathrm{FNN}$ in Eq.~\eqref{eq: fnn approximate f, in guarantee of fnn}. For any input $\bm{x} \in \mathbb{R}^I$, the output of the $\mathrm{F}\text{-}\mathrm{FTNet}_{\times}$ is 
\begin{equation*}
    \begin{aligned}
        f_{\times,\mathrm{F}}(\bm{x}) 
        =&~ \bm{\alpha}_{\times}^\top \mathrm{Re} \left[ \sigma_{\mathrm{\times}} \left( \left( \mathbf{W}_{\mathrm{\times}} + \mathbf{V}_{\mathrm{\times}} \mathrm{i} \right) \left( \kappa(\bm{x} , H_\times) + \bm{r}_0 \mathrm{i} \right) \right) \right] \\ 
        =&~ \bm{\alpha}_{\times}^\top \mathrm{Re} \left[ \sigma_{\times} \left( \left[ \left( \mathbf{W}_{\mathrm{F}} \bm{x} + \bm{b}_{\mathrm{F}} \right) + \mathbf{1} \mathrm{i} ; \mathbf{0} \right] \right) \right] \\ 
        =&~ \bm{\alpha}_{\mathrm{F}}^\top \mathrm{Re} \left[ \sigma_{\times} \left( \mathbf{W}_{\mathrm{F}} \bm{x} + \bm{b}_{\mathrm{F}} + \mathbf{1} \mathrm{i} \right) \right] , 
    \end{aligned}
\end{equation*}
where the first equality holds based on Eq.~\eqref{eq: mapping of f-ftnet}, the second and third equalities hold from the construction of $\mathrm{F}\text{-}\mathrm{FTNet}_{\times}$ in Step 1. Recalling the definition of the $\mathrm{zReLU}$ activation function in Eq.~\eqref{eq: definition of zrelu}, one has 
\begin{equation}
    \label{eq: mulftnet equiv to fnn, in guarantee of fnn}
    \begin{aligned}
        f_{\times,\mathrm{F}}(\bm{x}) 
        =&~ \bm{\alpha}_{\mathrm{F}}^\top \mathrm{Re} \left[ \left( \mathbf{W}_{\mathrm{F}} \bm{x} + \bm{b}_{\mathrm{F}} + \mathbf{1} \mathrm{i} \right) \circ \mathbb{I} \left( \mathbf{W}_{\mathrm{F}} \bm{x} + \bm{b}_{\mathrm{F}} \geqslant 0 \right) \right] \\ 
        =&~ \bm{\alpha}_{\mathrm{F}}^\top \left[ \left( \mathbf{W}_{\mathrm{F}} \bm{x} + \bm{b}_{\mathrm{F}} \right) \circ \mathbb{I} \left( \mathbf{W}_{\mathrm{F}} \bm{x} + \bm{b}_{\mathrm{F}} \geqslant 0 \right) \right] \\ 
        =&~ \bm{\alpha}_{\mathrm{F}}^\top \sigma_{\mathrm{F}} \left( \mathbf{W}_{\mathrm{F}} \bm{x} + \bm{b}_{\mathrm{F}} \right) \\ 
        =&~ f_{\mathrm{F}}(\bm{x}) , 
    \end{aligned}
\end{equation}
where the third equality holds because $\sigma_\mathrm{F}$ is the $\mathrm{ReLU}$ activation, and the fourth equality holds from Eq.~\eqref{eq: definition of fnn}. Finally, we prove that the constructed $\mathrm{F}\text{-}\mathrm{FTNet}_{\times}$ can $(\varepsilon,\mathcal{D})$-approximate the function $f$. According to Eqs.~\eqref{eq: fnn approximate f, in guarantee of fnn}~and~\eqref{eq: mulftnet equiv to fnn, in guarantee of fnn}, one has 
\begin{equation*}
    \mathbb{E}_{\bm{x}} \left[ ( f(\bm{x}) - f_{\times,\mathrm{F}}(\bm{x}) )^2 \right] 
    = \mathbb{E}_{\bm{x}} \left[ ( f(\bm{x}) - f_\mathrm{F}(\bm{x}) )^2 \right] 
    \leqslant \varepsilon , 
\end{equation*}
which completes the proof. $\hfill \square$ 

\noindent \textbf{Proof of Theorem~\ref{thm: worst case guarantee of ftnet over rnn}.} Since the target $\mathrm{DODS}$ $f_\mathrm{D}$ can be $(\varepsilon,\mathcal{D})$-approximated by $\mathrm{RNN}$, there exists an $\mathrm{RNN}$ defined by Eq.~\eqref{eq: definition of rnn} satisfying the following inequality
\begin{equation}
    \label{eq: rnn approximate dods, in guarantee of rnn}
    \mathbb{E}_{\bm{x}_{1:T} \sim \mathcal{D}} \left[ ( f_\mathrm{D}(\bm{x}_{1:T}) - f_\mathrm{R}(\bm{x}_{1:T}) )^2 \right] 
    \leqslant \varepsilon . 
\end{equation}

Firstly, we construct an $\mathrm{R}\text{-}\mathrm{FTNet}_{\times}$ $f_{\times,\mathrm{R}}$ with hidden size $H_{\times} = 2 H_{\mathrm{R}} + I + 1$ using the $\mathrm{zReLU}$ activation as follows 
\begin{equation}
    \label{eq: construction of mulftnet, in guarantee of rnn}
    \begin{aligned}
        & \mathbf{W}_{\times} = \left[ 
            \begin{array}{cccc}
                \mathbf{0}_{I \times I} & \mathbf{0} & \mathbf{0} & \mathbf{0} \\ 
                \mathbf{W}_{\mathrm{R}} & \mathbf{0}_{H_{\mathrm{R}} \times H_{\mathrm{R}}} & \mathbf{0} & \bm{\theta}_{\mathrm{R}} \\ 
                \mathbf{0} & \mathbf{0} & \mathbf{V}_{\mathrm{R}} & \mathbf{1} \\ 
                \mathbf{0} & \mathbf{0} & \mathbf{0} & \mathbf{0}_{1 \times 1} 
            \end{array}
        \right] ,
        \quad
        \mathbf{V}_{\times} = \left[ 
            \begin{array}{cccc}
                \mathbf{0} & \mathbf{0} & \mathbf{0} & \mathbf{0} \\ 
                \mathbf{0} & \mathbf{0} & - \mathbf{V}_{\mathrm{R}} & \mathbf{1} \\ 
                \mathbf{W}_{\mathrm{R}} & \mathbf{0} & \mathbf{0} & \bm{\theta}_{\mathrm{R}} \\ 
                \mathbf{0} & \mathbf{0} & \mathbf{0} & \mathbf{0} 
            \end{array}
        \right] , \\ 
        & \bm{r}_0 = \left[ \mathbf{0}_{I \times 1} ; \mathbf{0}_{H_{\mathrm{R}} \times 1} ; \bm{m}_0 ; \mathbf{0}_{1 \times 1} \right] ,
        \quad
        \bm{\alpha}_{\times} = \left[ \mathbf{0}_{I \times 1} ; \bm{\alpha}_{\mathrm{R}} ; \mathbf{0}_{H_{\mathrm{R}} \times 1} ; \mathbf{0}_{1 \times 1} \right] .
    \end{aligned}
\end{equation}

Secondly, we prove that the output of the constructed $\mathrm{R}\text{-}\mathrm{FTNet}_{\times}$ in Eq.~\eqref{eq: construction of mulftnet, in guarantee of rnn} is the same as that of the $\mathrm{RNN}$ in Eq.~\eqref{eq: rnn approximate dods, in guarantee of rnn}. Let $\bm{r}_t = [ \bm{r}_{t,1} ; \bm{r}_{t,2} ; \bm{r}_{t,3} ; \bm{r}_{t,4} ]$, where $\bm{r}_{t,1} \in \mathbb{R}^{I}$, $\bm{r}_{t,2} \in \mathbb{R}^{H_{\mathrm{R}}}$, $\bm{r}_{t,3} \in \mathbb{R}^{H_{\mathrm{R}}}$, and $\bm{r}_{t,4} \in \mathbb{R}$. We prove that $\bm{r}_{t,1} = \mathbf{0}_{I \times 1}$, $\bm{r}_{t,3} = \bm{m}_t$, and $\bm{r}_{t,4} = \mathbf{0}_{1 \times 1}$ hold for any $t \leqslant T$ by mathematical induction. 
\begin{enumerate}
    \item Base case. From Eq.~\eqref{eq: construction of mulftnet, in guarantee of rnn}, the claim holds for $t=0$. 
    \item Induction. Suppose that the claim holds for $t=\tau$ where $\tau \in \{0,1,\dots,T-1\}$. From Eq.~\eqref{eq: construction of mulftnet, in guarantee of rnn}, it is observed that the first and fourth rows of the weight matrices $\mathbf{W}_\times$ and $\mathbf{V}_\times$ are all $0$. Based on $\sigma_\times(0) = 0$, one knows that 
    \begin{equation*}
            \bm{r}_{\tau+1,1} = \sigma_\times(\bm{0}) = \bm{0} 
            \quad \text{and} \quad 
            \bm{r}_{\tau+1,4} = \sigma_\times(\bm{0}) = \bm{0} . 
    \end{equation*}
    Furthermore, one has 
    \begin{equation*}
        \begin{aligned}
            \bm{r}_{\tau+1,3} 
            =&~ \mathrm{Im} \left[ \sigma_{\times} \left( \mathbf{1} + \left( \mathbf{W}_{\mathrm{R}} \bm{x}_{\tau+1} + \mathbf{V}_{\mathrm{R}} \bm{r}_{\tau,3} + \bm{\theta}_{\mathrm{R}} \right) \mathrm{i} \right) \right] \\ 
            =&~ \left( \mathbf{W}_{\mathrm{R}} \bm{x}_{\tau+1} + \mathbf{V}_{\mathrm{R}} \bm{r}_{\tau,3} + \bm{\theta}_{\mathrm{R}} \right) \circ \mathbb{I} \left( \mathbf{W}_{\mathrm{R}} \bm{x}_{\tau+1} + \mathbf{V}_{\mathrm{R}} \bm{r}_{\tau,3} + \bm{\theta}_{\mathrm{R}} \geqslant 0 \right) \\ 
            =&~ \sigma_{\mathrm{R}} \left( \mathbf{W}_{\mathrm{R}} \bm{x}_{\tau+1} + \mathbf{V}_{\mathrm{R}} \bm{m}_{\tau} + \bm{\theta}_{\mathrm{R}} \right) \\ 
            =&~ \bm{m}_{\tau+1} , 
        \end{aligned}
    \end{equation*}
    where the first equality holds from the construction in Eq.~\eqref{eq: construction of mulftnet, in guarantee of rnn} and the hypothesis induction, the second equality holds according to Eq.~\eqref{eq: definition of zrelu}, the third equality holds because $\sigma_\mathrm{R}$ is the $\mathrm{ReLU}$ activation function, and the fourth equality holds based on Eq.~\eqref{eq: definition of rnn}. Thus, the claim holds for $t=\tau+1$. 
\end{enumerate}
For any $t \in [T]$, let $\bm{s}_t = [ \bm{s}_{t,1} ; \bm{s}_{t,2} ; \bm{s}_{t,3} ; \bm{s}_{t,4} ]$, where $\bm{s}_{t,1} \in \mathbb{R}^{I}$, $\bm{s}_{t,2} \in \mathbb{R}^{H_{\mathrm{R}}}$, $\bm{s}_{t,3} \in \mathbb{R}^{H_{\mathrm{R}}}$, and $\bm{s}_{t,4} \in \mathbb{R}$. Similar to the calculation of $\bm{r}_{t,3}$, one has $\bm{s}_{t,2} = \bm{m}_t$. Thus, the output of $\mathrm{R}\text{-}\mathrm{FTNet}_{\times}$ is the same as that of $\mathrm{RNN}$, i.e., 
\begin{equation}
    \label{eq: mulftnet equiv to rnn, in guarantee of rnn}
    y_{\times,t} 
    = \bm{\alpha}_{\times}^\top \bm{s}_t 
    = \bm{\alpha}_{\mathrm{R}}^\top \bm{s}_{t,2} 
    = \bm{\alpha}_{\mathrm{R}}^\top \bm{m}_{t} 
    = y_{\mathrm{R},t} . 
\end{equation}

Finally, we prove that the constructed $\mathrm{R}\text{-}\mathrm{FTNet}_{\times}$ can $(\varepsilon,\mathcal{D})$-approximate $f_\mathrm{D}$. According to Eqs.~\eqref{eq: rnn approximate dods, in guarantee of rnn}~and~\eqref{eq: mulftnet equiv to rnn, in guarantee of rnn}, one has 
\begin{equation*}
    \mathbb{E}_{\bm{x}_{1:T} \sim \mathcal{D}} \left[ ( f_\mathrm{D}(\bm{x}_{1:T}) - f_{\times,\mathrm{R}}(\bm{x}_{1:T}) )^2 \right]
    = \mathbb{E}_{\bm{x}_{1:T} \sim \mathcal{D}} \left[ ( f_\mathrm{D}(\bm{x}_{1:T}) - f_\mathrm{R}(\bm{x}_{1:T}) )^2 \right] 
    \leqslant \varepsilon , 
\end{equation*}
which completes the proof. $\hfill \square$

\begin{table}[H]
    \renewcommand\arraystretch{1.5}
    \centering
    \caption{Approximation Complexity of $\mathrm{FTNet}$ and $\mathrm{FNN}$}
    \label{tab: approximation complexity of ftnet and fnn}
    \begin{tabular}{ccc}
        \hline
        Target & Width of $\mathrm{FNN}$ & Width of $\mathrm{FTNet}$ \\
        \hline
        In Theorem~\ref{thm: advantage of ftnet over fnn} & $\Omega(\mathrm{e}^{\epsilon_1 I} / I)$ & $O(I^{15/4})$ \\
        \hline
        Any (Theorem~\ref{thm: worst case guarantee of ftnet over fnn}) & $H_F$ & $O(H_F)$ \\
        \hline
    \end{tabular}
\end{table}

\begin{table}[H]
    \renewcommand\arraystretch{1.5}
    \centering
    \caption{Approximation Complexity of $\mathrm{FTNet}$ and $\mathrm{RNN}$}
    \label{tab: approximation complexity of ftnet and rnn}
    \begin{tabular}{ccc}
        \hline
        Target & Width of $\mathrm{RNN}$ & Width of $\mathrm{FTNet}$ \\
        \hline
        In Theorem~\ref{thm: advantage of ftnet over rnn} & $\Omega(\mathrm{e}^{\varepsilon_2 I})$ & $O(I^{15/4})$ \\
        \hline
        Any (Theorem~\ref{thm: worst case guarantee of ftnet over rnn}) & $H_R$ & $O(H_R)$ \\
        \hline
    \end{tabular}
\end{table}

Tables~\ref{tab: approximation complexity of ftnet and fnn}~and~\ref{tab: approximation complexity of ftnet and rnn} summarize the approximation complexity results of $\mathrm{FTNet}$ using asymptotic notations, where $\epsilon_1$ and $\epsilon_2$ are two constants irrelevant to the input dimension $I$. $\mathrm{FTNet}$ possesses exponential advantage when expressing particular functions, and requires hidden size of the same order in arbitrary cases. These results suggest that $\mathrm{FTNet}$ is able to exhibit dynamic reaction by the flexible formulation of the synapse, which would be demanded in decision making~\cite{zhou2022rehearsal} and open-environment machine learning~\cite{zhou2022open}, though the analysis is beyond the scope of this paper.

\section{Local Minima}
\label{sec: local minima}

This section investigates the empirical loss surface of $\mathrm{F}\text{-}\mathrm{FTNet}$. Let $S = \{ (\bm{x}_i,y_i) \}_{i=1}^n$ be the training set, where $\bm{x}_i \in \mathbb{R}^I$ denotes the $i$-th sample, and $y_i \in \mathbb{R}$ represents the label of the $i$-th sample. Consider the empirical loss of $\mathrm{F}\text{-}\mathrm{FTNet}$ with the following form 
\begin{equation}
    \label{eq: definition of empirical loss}
    \hat{L} = \sum_{i=1}^n l \left( f_{\times,\mathrm{F}}(\bm{x}_i) - y_i \right) , 
\end{equation}
where $f_{\times,\mathrm{F}}$ is the mapping of $\mathrm{F}\text{-}\mathrm{FTNet}$ defined in Eq.~\eqref{eq: mapping of f-ftnet}, and $l : \mathbb{R} \rightarrow \mathbb{R}$ is a loss function. Let $\mathbf{Z} = \mathbf{W}_\times + \mathbf{V}_\times \mathrm{i}$ be the complex-valued weight matrix, and $\bm{\alpha}$ denotes $\bm{\alpha}_\times$ for simplicity. Then the empirical loss $\hat{L}$ is a function of $\mathbf{Z}$ and $\bm{\alpha}$, denoted by $\hat{L} ( \mathbf{Z} , \bm{\alpha} )$. Holomorphic activation functions are of interest in this section, and the definition of holomorphic functions is reviewed as follows. 

\begin{definition}{\cite[Page 2]{gunning2009analytic}}
    A function $g : \mathbb{C}^m \rightarrow \mathbb{C}$ is called holomorphic if for each point $\bm{w} = (w_1,w_2,\dots,w_m) \in \mathbb{C}^m$, there exists an open set $U$, such that $\bm{w} \in U$, and the function $g$ has a power series expansion 
    \begin{equation}
        \label{eq: definition of holomorphic}
        f(\bm{z}) = \sum_{(v_1,v_2,\dots,v_m) \in \mathbb{N}^m} a_{v_1,v_2,\dots,v_m} \prod_{j=1}^m (z_j-w_j)^{v_j} , 
    \end{equation}
    which converges for all $\bm{z} = (z_1,z_2,\dots,z_m) \in U$. 
\end{definition}

Let us define a class of loss functions called \emph{well-posed regression loss functions}. 

\begin{definition}
    \label{def: well-posed regression loss functions}
    A loss function $l : \mathbb{R} \rightarrow \mathbb{R}$ is called a well-posed regression loss function, if $l$ satisfies the following conditions: i) $l$ is analytic on $\mathbb{R}$; ii) $l(0) = 0$; iii) $l$ is strictly decreasing on $(-\infty,0)$ and strictly increasing on $(0,+\infty)$. 
\end{definition}

The conditions in Definition~\ref{def: well-posed regression loss functions} are satisfied by many commonly used loss functions for regression or their smooth variants, such as the squared loss $l(x) = x^2$, the parameterized $\mathrm{cosh}$ $l(x) = c^{-1} [ \ln ( \mathrm{e}^{ax} + \mathrm{e}^{-bx} ) - \ln 2 ]$ with positive parameters $a$, $b$, and $c$, which can approximate the absolute loss $l(x) = |x|$ and the quantile loss $l(x) = (1-\theta) x \mathbb{I}_{x \geqslant 0} - \theta x \mathbb{I}_{x < 0}$ with $\theta \in (0,1)$~\cite{koenker1978regression} in the limit. The following theorem studies local minima of the empirical loss $\hat{L}$. 

\begin{theorem}
    \label{thm: local minima of f-ftnet has loss 0}
    Suppose that all samples are linearly independent, the activation function $\sigma_\times$ is holomorphic and not polynomial, and that $l$ is a well-posed regression loss function. If the loss $\hat{L} ( \mathbf{Z} , \bm{\alpha} )$ is positive, then for any $\delta > 0$, there exist $\Delta \mathbf{Z}$ and $\Delta \bm{\alpha}$ satisfying the following inequalities
    \begin{equation*}
        \hat{L} \left( \mathbf{Z} + \Delta \mathbf{Z} , \bm{\alpha} + \Delta \bm{\alpha} \right) 
        < \hat{L} \left( \mathbf{Z} , \bm{\alpha} \right)
    \end{equation*}
    and
    \begin{equation*}
        \left\Vert \Delta \mathbf{Z} \right\Vert_F + \left\Vert \Delta \bm{\alpha} \right\Vert_2 \leqslant \delta .
    \end{equation*}
\end{theorem}

Theorem~\ref{thm: local minima of f-ftnet has loss 0} shows that it is always possible to reduce the loss in the neighborhood as long as the loss is not $0$. This indicates that any local minimum of $\hat{L} ( \mathbf{Z} , \bm{\alpha} )$ is the global minimum. The requirement of linearly independent samples holds with probability $1$ when the sample size is no larger than the input dimension, and the samples are generated from a continuous distribution. Existing studies mostly investigate the local-minima-free condition of FNN using specific loss with strong conditions, such as linearly separable data~\cite{liang2018understanding}, particular activation functions~\cite{kawaguchi2016deep,soltanolkotabi2017learning,liang2018adding,laurent2018deep}, and over-parameterization together with special initialization~\cite{allen2019convergence2,du2019gradient,jacot2018neural,zou2019improved,zou2020gradient}. Theorem~\ref{thm: local minima of f-ftnet has loss 0} holds for a large class of activations and loss functions. The requirement of a large input dimension is reasonable since previous work proves that FNN has suboptimal local minima with low-dimensional input under general settings~\cite{ding2022suboptimal}. We begin our proof with two lemmas.

\begin{lemma}
    \label{lem: neighborhood of holomorphic functions}
    Let $g : \mathbb{C}^m \rightarrow \mathbb{C}$ be holomorphic and not constant. For any $\bm{z}^{(0)} \in \mathbb{C}^m$, $\delta \in (0,1)$, there exist $\Delta \bm{z}^{(1)} , \Delta \bm{z}^{(2)} \in \mathbb{C}^m$ satisfying the following inequalities
    \begin{eqnarray*}
        \left\Vert \Delta \bm{z}^{(1)} \right\Vert_2^2 \leqslant \delta 
        \quad \text{and} \quad 
        \mathrm{Re} \left[ g(\bm{z}^{(0)} + \Delta \bm{z}^{(1)}) \right] > \mathrm{Re} \left[ g(\bm{z}^{(0)}) \right] , \\ 
        \left\Vert \Delta \bm{z}^{(2)} \right\Vert_2^2 \leqslant \delta 
        \quad \text{and} \quad 
        \mathrm{Re} \left[ g(\bm{z}^{(0)} + \Delta \bm{z}^{(2)}) \right] < \mathrm{Re} \left[ g(\bm{z}^{(0)}) \right] . 
    \end{eqnarray*}
\end{lemma}

Lemma~\ref{lem: neighborhood of holomorphic functions} shows that the neighborhood of holomorphic functions possesses rich diversity, i.e., one can always find a point with a smaller real part and another point with a larger real part in the neighborhood. 

\noindent \textbf{Proof of Lemma~\ref{lem: neighborhood of holomorphic functions}.} Let $\bm{z}^{(0)} = ( z_1^{(0)} , z_2^{(0)} , \dots , z_m^{(0)} )$. Since the function $g$ is holomorphic, by rearranging Eq.~\eqref{eq: definition of holomorphic}, there exist an open set $U$ and a series of holomorphic functions $f_j^{(k)} : \mathbb{C}^{m-k} \rightarrow \mathbb{C}$ with $j \in \mathbb{N}$ and $k \in [m]$, s.t. $\bm{z}^{(0)} \in U$, and for any $\bm{z} = (z_1,z_2,\dots,z_m) \in U$, the following holds
\begin{equation*}
  \begin{aligned}
    & f_0^{(0)}(\bm{z}) := g(\bm{z}) = \sum_{j=0}^\infty f^{(1)}_j (z_2,z_3,\dots,z_m) \left( z_1 - z_1^{(0)} \right)^j , \\ 
    & f^{(1)}_0 (z_2,z_3,\dots,z_m) = \sum_{j=0}^\infty f^{(2)}_j (z_3,z_4,\dots,z_m) \left( z_2 - z_2^{(0)} \right)^j , \\ 
    & f^{(2)}_0 (z_3,z_4,\dots,z_m) = \sum_{j=0}^\infty f^{(3)}_j (z_4,z_5,\dots,z_m) \left( z_3 - z_3^{(0)} \right)^j , \\ 
    & \dots \dots \\ 
    & f^{(m-1)}_0 (z_m) = \sum_{j=0}^\infty f^{(m)}_j \left( z_m - z_m^{(0)} \right)^j . 
  \end{aligned}
\end{equation*}
It is observed that $f_0^{(m)} = g(\bm{z}^{(0)})$ when $\bm{z} = \bm{z}^{(0)}$. Then the following $k_0$ is well-defined 
\begin{equation*}
    k_0 = \min \left\{ k \in [m] ~\bigg|~ f_0^{(k)} \equiv g\left(\bm{z}^{(0)}\right) \right\} . 
\end{equation*}
Let $\bm{z}^{(1)} = ( z_1^{(0)} , \dots , z_{k_0-1}^{(0)} , z_{k_0} , \dots , z_m )$. Thus, one has 
\begin{equation*}
    g\left(\bm{z}^{(1)}\right) 
    = g\left(\bm{z}^{(0)}\right) + \sum_{j=1}^\infty f_j^{(k_0)}(z_{k_0+1},\dots,z_m) \left( z_{k_0} - z_{k_0}^{(0)} \right)^j . 
\end{equation*}
Since $f_0^{(k_0-1)} \not \equiv g( \bm{z}^{(0)} )$ and $g(\bm{z}^{(1)}) = f_0^{(k_0-1)}$, one has $g(\bm{z}^{(1)}) \not \equiv g( \bm{z}^{(0)} )$. Thus, there exists a positive integer $j \in \mathbb{N}^+$, such that $f_j^{(k_0)}(z_{k_0+1},z_{k_0+2},\dots,z_m) \not \equiv 0$. Then the following $j_0$ is well-defined 
\begin{equation*}
    j_0 = \min \left\{ j \in \mathbb{N}^+ ~\bigg|~ f_j^{(k_0)}(z_{k_0+1},z_{k_0+2},\dots,z_m) \not \equiv 0 \right\} . 
\end{equation*}
Therefore, there exist $z_{k_0+1}^{(1)} , z_{k_0+2}^{(1)} , \dots , z_m^{(1)}$, such that 
\begin{equation*}
    f_{j_0}^{(k_0)} \left( z_{k_0+1}^{(1)} , z_{k_0+2}^{(1)} , \dots , z_m^{(1)} \right) \neq 0 ,
\end{equation*}
and
\begin{equation}
    \label{eq: difference is small, in proof of neighbor of holomorphic}
    \sum_{k=k_0+1}^{m} \left( z_k^{(1)} - z_k^{(0)} \right)^2 \leqslant \frac{\delta}{2} .
\end{equation}
Let $\bm{z}^{(2)} = ( z_1^{(0)} , z_2^{(0)} , \dots , z_{k_0-1}^{(0)} , z_{k_0} , z_{k_0+1}^{(1)} , z_{k_0+2}^{(1)} , \dots , z_m^{(1)} )$. Then the function value of $g$ at $\bm{z}^{(2)}$ satisfies
\begin{equation}
    \label{eq: series, in proof of neighbor of holomorphic}
    g\left(\bm{z}^{(2)}\right) 
    = g\left(\bm{z}^{(0)}\right) + \sum_{j=j_0}^\infty a_j \left( z_{k_0} - z_{k_0}^{(0)} \right)^j , 
\end{equation}
where $a_j = f_j^{(k_0)} ( z_{k_0+1}^{(1)} , z_{k_0+2}^{(1)} , \dots , z_m^{(1)} )$ and $a_{j_0} \neq 0$. Since $\bm{z}^{(0)}$ is in the open set $U$, there exists $r > 0$, such that the ball $B( \bm{z}^{(0)} , r )$ is a subset of $U$. Then the radius of convergence of the series in Eq.~\eqref{eq: series, in proof of neighbor of holomorphic} is at least $r$. Thus, one has $\lim \sup_{j \rightarrow \infty} |a_j|^{1/j} \leqslant 1/r$ from the Cauchy-Hadamard theorem. Since any series with finite limit superior is bounded, there exists $M \geqslant \max \{ 1 , \sqrt{2\delta}/3 \}$, such that $|a_j|^{1/j} \leqslant M$, i.e., $|a_j| \leqslant M^j$. Define the change of $\bm{z}^{(0)}$ as
\begin{equation*}
    \Delta \bm{z}^{(1)} = \left( 0 , \dots , 0 , \tilde{z}_{k_0} , z_{k_0+1}^{(1)} - z_{k_0+1}^{(0)} , \dots , z_m^{(1)} - z_m^{(0)} \right) , 
\end{equation*}
where 
\begin{equation*}
  \tilde{z}_{k_0} = \frac{ \min \left\{ 1 , |a_{j_0}| \right\} }{3M^{j_0+1} \sqrt{2/\delta}} \mathrm{e}^{ - \mathrm{i} \theta_{a_{j_0}} / j_0 } . 
\end{equation*}
In view of $M \geqslant 1$ and Eq.~\eqref{eq: difference is small, in proof of neighbor of holomorphic}, one has 
\begin{equation*}
    \left\Vert \Delta \bm{z}^{(1)} \right\Vert_2^2 
    \leqslant \left| \tilde{z}_{k_0} \right|^2 + \sum_{k=k_0+1}^m \left( z_k^{(1)} - z_k^{(0)} \right)^2 
    \leqslant \delta , 
\end{equation*}
meanwhile,
\begin{equation*}
    \begin{aligned}
        \mathrm{Re} \left[ g \left( \bm{z}^{(0)} + \Delta \bm{z}^{(1)} \right) \right] - \mathrm{Re} \left[ g \left( \bm{z}^{(0)} \right) \right]
        &= \sum_{j=j_0}^\infty \mathrm{Re} \left[ a_j \left( \tilde{z}_{k_0} \right)^j \right] \\ 
        &\geqslant |a_{j_0}| \left| \tilde{z}_{k_0} \right|^{j_0} - \sum_{j=j_0+1}^\infty M^j \left| \tilde{z}_{k_0} \right|^j \\ 
        &\geqslant \min \left\{ 1 , |a_{j_0}| \right\}^{j_0+1} \frac{ (\delta/2)^{j_0/2} }{ 3^{j_0} M^{j_0(j_0+1)} } \left[ 1 - \frac{ 2 }{3 \sqrt{2/\delta}} \right] \\ 
        &> 0 , 
    \end{aligned}
\end{equation*}
where the first equality holds from Eq.~\eqref{eq: series, in proof of neighbor of holomorphic}, the first inequality holds because of $\mathrm{Re}[z] \geqslant -|z|$ and $|a_j| \leqslant M^j$, the second inequality holds based on $M \geqslant \sqrt{2\delta} / 3$, and the third inequality holds in view of $\delta < 1$. Let 
\begin{equation*}
    \Delta \bm{z}^{(2)} = \left( 0 , \dots , 0 , \hat{z}_{k_0} , z_{k_0+1}^{(1)} - z_{k_0+1}^{(0)} , \dots , z_m^{(1)} - z_m^{(0)} \right) , 
\end{equation*}
where 
\begin{equation*}
  \hat{z}_{k_0} = \frac{ \min \left\{ 1 , |a_{j_0}| \right\} }{3M^{j_0+1} \sqrt{2/\delta}} \mathrm{e}^{ - \mathrm{i} \left( \pi + \theta_{a_{j_0}} \right) / j_0 } . 
\end{equation*}
Then the conclusion about $\Delta \bm{z}^{(2)}$ can be proven similarly. $\hfill \square$ 

\begin{lemma}
    \label{lem: not a constant} 
    Let $\sigma : \mathbb{C} \rightarrow \mathbb{C}$ be holomorphic and not polynomial, $\left\{ \bm{x}^{(j)} \right\}_{j=1}^n \subset \mathbb{R}^m$ are $n$ different vectors, $\left\{ y_j \right\}_{j=1}^n \subset \mathbb{R}$ are not all zero, and $\bm{z} = (z_1,z_2,\dots,z_{m+1})$ is a complex-valued vector. Then the function $ g : \mathbb{C}^{m+1} \rightarrow \mathbb{C}$, defined by  
    \begin{equation*}
        g(\bm{z}) = \sum_{j=1}^n y_j \sigma \left( x_1^{(j)} z_1 + x_2^{(j)} z_2 + \dots + x_m^{(j)} z_m + z_{m+1} \right) , 
    \end{equation*}
    is not a constant function. 
\end{lemma}

Lemma~\ref{lem: not a constant} provides a sufficient condition that the summation of the activation of weighted average is not a constant. 

\noindent \textbf{Proof of Lemma~\ref{lem: not a constant}.} The proof consists of several steps. \\ 
\textbf{Step 1.} We find the necessary condition that $g$ is a constant. Since $\sigma$ is holomorphic, and any holomorphic function coincides with its Taylor series in any open set within the domain of the function~\cite[Theorem 4.4]{stein2010complex}, there exists $\{ c_k \}_{k=0}^\infty \subset \mathbb{C}$, such that $\sigma(z) = \sum_{k=0}^\infty c_k z^k$ holds for any $z \in \mathbb{C}$. Since $\sigma$ is not polynomial, there exists $\{ n_k \}_{k=1}^\infty \subset \mathbb{N}^+$, such that $\sigma(z) = c_0 + \sum_{k=1}^\infty c_{n_k} z^{n_k}$, where $n_k < n_{k+1}$ and $c_{n_k} \neq 0$ hold for any $k \in \mathbb{N}^+$. Thus, the function $g$ can be rewritten as 
\begin{equation*}
    g(\bm{z}) 
    = \sum_{j=1}^n y_j \left[ c_0 + \sum_{k=1}^\infty c_{n_k} \left( z_{m+1} + \sum_{l=1}^{m} x_l^{(j)} z_l \right)^{n_k} \right]
    = h_0(\bm{z}) + \sum_{k=1}^\infty h_{n_k}(\bm{z}) . 
\end{equation*}
If $g$ is a constant, then one has 
\begin{equation*}
    \sum_{j=1}^n y_j \left( z_{m+1} + \sum_{l=1}^m x_l^{(j)} z_l \right)^{n_k} \equiv 0 , 
    \quad \forall ~ k \in \mathbb{N}^+ . 
\end{equation*}
According to the multinomial theorem, one has 
\begin{equation*}
    \sum_{j=1}^n \sum_{\bm{p} \in P_{n_k}} y_j c_{\bm{p}} z_{m+1}^{p_{m+1}} \prod_{l=1}^m \Big( x_l^{(j)} z_l \Big)^{p_l} \equiv 0 , 
    \quad \forall ~ k \in \mathbb{N}^+ , 
\end{equation*}
where $\bm{p} = (p_1,p_2,\dots,p_{m+1})$, $P_{n_k} = \{ \bm{p} \mid \forall ~ l \in [m+1] , p_l \in \mathbb{N} , \Vert \bm{p} \Vert_1 = n_k \}$, and $c_{\bm{p}}$ is the multinomial coefficient. Since $z_1,z_2,\dots,z_{m+1}$ are free variables, one has 
\begin{equation*}
    \sum_{j=1}^n y_j \prod_{l=1}^m \Big( x_l^{(j)} \Big)^{p_l} = 0 , \quad \forall ~ k \in \mathbb{N}^+ , \bm{p} \in P_{n_k} . 
\end{equation*}
Since $n_k \rightarrow + \infty$ as $k \rightarrow + \infty$, we obtain the following necessary condition of constant $g$
\begin{equation}
    \label{eq: necessary condition, in proof of not constant}
    \sum_{j=1}^n y_j \prod_{l=1}^m \Big( x_l^{(j)} \Big)^{p_l} = 0 , \quad \forall ~ p_1 , p_2 , \dots , p_m \in \mathbb{N} . 
\end{equation}
\textbf{Step 2.} We restrict the summation domain of the necessary condition in Eq.~\eqref{eq: necessary condition, in proof of not constant} to obtain another necessary condition. Let $J_0 = \{ j \in [n] \mid y_j \neq 0 \}$. For any $l \in [m]$, we define 
\begin{equation*}
    \begin{aligned}
        m_l = \max_{j \in J_{l-1}} \left| x_l^{(j)} \right| 
        \quad \text{and} \quad 
        J_l = \left\{ j \in J_{l-1} ~\bigg|~ \left| x_l^{(j)} \right| = m_l \right\} . 
    \end{aligned}
\end{equation*}
Under the assumption that $g$ is constant, we claim that 
\begin{equation}
    \label{eq: restricted sum is 0, in proof of not constant}
    \sum_{j \in J_m} y_j \prod_{l=1}^m \Big( x_l^{(j)} \Big)^{p_l} = 0 , \quad \forall ~ p_1 , p_2 , \dots , p_m \in \mathbb{N} . 
\end{equation}
Otherwise, there exist $q_1 , q_2 , \dots , q_m \in \mathbb{N}$, such that 
\begin{equation*}
    \sum_{j \in J_m} y_j \prod_{l=1}^m \Big( x_l^{(j)} \Big)^{q_l} = c_0 \neq 0 . 
\end{equation*}
Let $r_1,r_2,\dots,r_m$ be even natural numbers. Thus, one has 
\begin{equation*}
    0 
    = \left| \sum_{j \in J_0} y_j \prod_{l=1}^m \Big( x_l^{(j)} \Big)^{q_l+r_l} \right|
    \geqslant \left| \sum_{j \in J_m} y_j \prod_{l=1}^m \Big( x_l^{(j)} \Big)^{q_l+r_l} \right| - \sum_{l=1}^m \left| \sum_{j \in J_{l-1} \backslash J_l} y_j \prod_{l=1}^m \Big( x_l^{(j)} \Big)^{q_l+r_l} \right| , 
\end{equation*}
where the equality holds from Eq.~\eqref{eq: necessary condition, in proof of not constant} and definition of $J_0$, and the inequality holds based on the triangle inequality. Let 
\begin{equation*}
    y_M = \max_{j \in [n]} |y_j| 
    \quad \text{and} \quad 
    L = \left\{ l \in [m] ~\big|~ |J_{l-1} \backslash J_l| \geqslant 1 \right\} . 
\end{equation*}
For $l \in [m]$, define 
\begin{equation*}
	\begin{aligned}
		& M_l = \max_{j \in [n]} \left| x_l^{(j)} \right| , \\ 
        & m_{l,2} = \left\{ 
            \begin{aligned}
                & \max_{j \in J_{l-1} \backslash J_l} \left| x_l^{(j)} \right| , & \text{if}~ J_{l-1} \neq J_l , \\ 
                & 0 , & \text{if}~ J_{l-1} = J_l , 
            \end{aligned}
        \right. \\ 
        & \mathcal{A}_l = \left( \prod_{s=1}^{l-1} m_s^{q_s+r_s} \right) \left( \prod_{t=l+1}^m M_t^{q_t+r_t} \right) . 
    \end{aligned}
\end{equation*}
Thus, one has 
\begin{equation*}
    \begin{aligned}
        0 
        \geqslant&~ |c_0| \prod_{l=1}^m m_l^{r_l} - \sum_{l=1}^m \left|J_{l-1} \backslash J_l\right| y_M \mathcal{A}_l m_{l,2}^{r_l} \\
        \geqslant&~ |c_0| \prod_{l=1}^m m_l^{r_l} - \sum_{l \in L} \left|J_{l-1} \backslash J_l\right| y_M \mathcal{A}_l m_{l,2}^{r_l} \\ 
        \geqslant&~ |c_0| \prod_{l=1}^m m_l^{r_l} - n y_M \sum_{l \in L} \mathcal{A}_l m_{l,2}^{r_l} , 
    \end{aligned}
\end{equation*}
where the second inequality holds because of $J_{l-1} \backslash J_l \subset J_{l-1} \subset [n]$. For $l \in L$, it is observed that $m_l > m_{l,2} \geqslant 0$. Thus, the second term in the above inequality will be much smaller than the first term when $r_l$ is sufficiently large. More formally, we define $r_l$ as follows. \\ 
\textbf{Case 1.} If $l \in [n] \backslash L$, define $r_l = 0$. \\ 
\textbf{Case 2.} If $l \in L$ and $m_{l',2} = 0$, define $r_l = 2$. \\ 
\textbf{Case 3.} Otherwise, define $\lceil x \rceil_{\mathrm{E}}$ as the smallest even integer no less than $x$. Let 
\begin{equation*}
    r_{l'} = \left\lceil \frac{ 1 }{ \ln \left( \frac{ m_{l',2} }{ m_{l'} } \right)  } \ln \left( \frac{ |c_0| \prod_{t=l'+1}^m M_t^{q_t+r_t} }{ 2 n^2 y_M \prod_{s=1}^{l'-1} m_s^{q_s} \prod_{t=l'+1}^m m_t^{r_t} } \right) \right\rceil_{\mathrm{E}} . 
\end{equation*}
Based on the choice of $r_l$, one has 
\begin{equation*}
        0 
        \geqslant |c_0| \prod_{l=1}^m m_l^{r_l} - n y_M \sum_{l \in L} \frac{ |c_0| \prod_{l'=1}^m m_{l'}^{r_{l'}} }{ 2 n^2 y_M } 
        \geqslant \frac{|c_0|}{2} \prod_{l=1}^m m_l^{r_l} , 
\end{equation*}
where the second inequality holds because of $|L| \leqslant n$. When $m_l = 0$, one has $J_{l-1} = J_l$ from the definition of $J_l$. Thus, one has $l \in L$, which leads to $r_l = 0$. Since $c_0 \neq 0$, one has 
\begin{equation*}
    0 
    \geqslant \frac{|c_0|}{2} \prod_{l=1}^m m_l^{r_l} 
    > 0 , 
\end{equation*}
which is a contradiction. Thus, we have proven the claim in Eq.~\eqref{eq: restricted sum is 0, in proof of not constant}. It is observed that $| x_l^{(j)} | = m_l$ holds for any $j \in J_m$. Thus, the claim indicates that when $g$ is a constant, one has 
\begin{equation}
    \label{eq: restricted sum of sign is 0, in proof of not constant}
    \sum_{j \in J_m} y_j \prod_{l=1}^m \mathrm{sign} \Big( x_l^{(j)} \Big)^{p_l} = 0 , \quad \forall ~ p_1 , p_2 , \dots , p_m \in \mathbb{N} , 
\end{equation}
where $\mathrm{sign}(\cdot)$ denotes the sign function. \\ 
\textbf{Step 3.} We prove that the necessary condition of constant $g$ in Eq.~\eqref{eq: restricted sum of sign is 0, in proof of not constant} does not hold by probabilistic methods. For any $j \in J_m$, let $N_j = \{ l \in [m] \mid \bm{x}_l^{(j)} = - m_l \}$ denote the set of dimensions in which $\bm{x}^{(j)}$ is negative. Observe that $\{ N_j \}_{j \in J_m}$ are different since $\{ \bm{x}^{(j)} \}_{j=1}^n$ are different. Thus, there exists a minimal element among $\{ N_j \}_{j \in J_m}$, i.e., there exists $j_0 \in J_m$, such that for any $j \in S_m \backslash \{ j_0 \}$, one has $T_j \not \subset T_{j_0}$. For any $l \in [m]$, we define a random variable $\sigma_l$ as follows 
\begin{equation*}
    \left\{\begin{aligned}
        &\mathrm{Pr}[\sigma_l=0] = 1 , & \text{if}~ l \in T_{j_0} , \\ 
        & \mathrm{Pr}[\sigma_l=0] = \mathrm{Pr}[\sigma_l=1] = 1/2 , & \text{if}~ l \not \in T_{j_0} . 
    \end{aligned}\right.
\end{equation*}
Let $\bm{\sigma} = (\sigma_1,\sigma_2,\dots,\sigma_m)$. Thus, one has 
\begin{equation}
    \label{eq: temp 1, in proof of not constant}
    \begin{aligned}
        0 
        =&~ \mathbb{E}_{\bm{\sigma}} \left[ \sum_{j \in J_m} y_j \prod_{l=1}^m \mathrm{sign} \Big( x_l^{(j)} \Big)^{\sigma_l} \right] \\ 
        =&~ \mathbb{E}_{\bm{\sigma}} \left[ y_{j_0} \prod_{l=1}^m \mathrm{sign} \Big( x_l^{(j)} \Big)^{\sigma_l} \right] + \mathbb{E}_{\bm{\sigma}} \left[ \sum_{j \in J_m \backslash \{j_0\}} y_j \prod_{l=1}^m \mathrm{sign} \Big( x_l^{(j)} \Big)^{\sigma_l} \right] , 
    \end{aligned}
\end{equation}
where the first equality holds from Eq.~\eqref{eq: restricted sum of sign is 0, in proof of not constant}. For the first term of Eq.~\eqref{eq: temp 1, in proof of not constant}, the following equation holds
\begin{equation}
    \label{eq: term 1 of temp 1, in proof of not constant}
    \mathbb{E}_{\bm{\sigma}} \left[ y_{j_0} \prod_{l=1}^m \mathrm{sign} \Big( x_l^{(j)} \Big)^{\sigma_l} \right]
    = y_{j_0} \mathbb{E}_{\bm{\sigma}} \left[ \prod_{l \in T_{j_0}} \mathrm{sign} \Big( x_l^{(j_0)} \Big)^{\sigma_l} \prod_{l \not \in T_{j_0}} \mathrm{sign} \Big( x_l^{(j_0)} \Big)^{\sigma_l} \right]
    = y_{j_0} , 
\end{equation}
where the second equality holds because of $\sigma_l = 0$ for all $l \in T_{j_0}$ and $x_l^{(j_0)} > 0$ for all $l \not \in T_{j_0}$. Since $T_j \not \subset T_{j_0}$ holds for any $j \in J_m \backslash \{j_0\}$, there exists $l_j$ such that $l_j \not \in T_{j_0}$ and $l_j \in T_j$. Thus, for the second term of Eq.~\eqref{eq: temp 1, in proof of not constant}, one has 
\begin{equation}
    \label{eq: term 2 of temp 1, in proof of not constant}
    \begin{aligned}
        \mathbb{E}_{\bm{\sigma}} \left[ \sum_{j \in J_m \backslash \{j_0\}} y_j \prod_{l=1}^m \mathrm{sign} \Big( x_l^{(j)} \Big)^{\sigma_l} \right]
        &= \sum_{j \in J_m \backslash \{j_0\}} y_j \mathbb{E}_{\bm{\sigma}} \mathbb{E}_{\sigma_{l_j}} \left[ \prod_{l=1}^m \mathrm{sign} \Big( x_l^{(j)} \Big)^{\sigma_l} \right] \\ 
        &= \sum_{j \in J_m \backslash \{j_0\}} y_j \mathbb{E}_{\bm{\sigma}} \left[ \prod_{l=1,l \neq l_j}^m \mathrm{sign} \Big( x_l^{(j)} \Big)^{\sigma_l} \cdot 0 \right] \\ 
        &= 0 , 
    \end{aligned}
\end{equation}
where the second equality holds since $x_l^{(j)} < 0$ and $\mathrm{Pr}[\sigma_{l_j}=0] = \mathrm{Pr}[\sigma_{l_j}=1] = 1/2$. Substituting Eqs.~\eqref{eq: term 1 of temp 1, in proof of not constant} and~\eqref{eq: term 2 of temp 1, in proof of not constant} into Eq.~\eqref{eq: temp 1, in proof of not constant}, one has $y_{j_0} = 0$, which contradicts the fact that $j_0 \in J_m \subset J_0$ and $y_j \neq 0$ for all $j \in J_0$. Thus, the necessary condition of constant $g$ in Eq.~\eqref{eq: restricted sum of sign is 0, in proof of not constant} does not hold, which leads to the conclusion that $g$ is not a constant. $\hfill \square$ 

\noindent \textbf{Proof of Theorem~\ref{thm: local minima of f-ftnet has loss 0}.} Let $\Delta \hat{L} ( \Delta \mathbf{Z} , \Delta \bm{\alpha} ) = \hat{L} ( \mathbf{Z} + \Delta \mathbf{Z} , \bm{\alpha} + \Delta \bm{\alpha} ) - \hat{L} ( \mathbf{Z} , \bm{\alpha} ) $ denote the change of empirical loss. Recalling Eqs.~\eqref{eq: mapping of f-ftnet} and~\eqref{eq: definition of empirical loss}, one has 
\begin{equation*}
    \Delta \hat{L} ( \Delta \mathbf{Z} , \Delta \bm{\alpha} ) 
    = \sum_{j=1}^n - l \left( \bm{\alpha}^\top \mathrm{Re} \left[ \sigma \left( \mathbf{Z} \bm{\kappa}_j \right) \right] - y_j \right) + l \left( \left( \bm{\alpha} + \Delta \bm{\alpha} \right)^\top \mathrm{Re} \left[ \sigma \left( \left( \mathbf{Z} + \Delta \mathbf{Z} \right) \bm{\kappa}_j \right) \right] - y_j \right) , 
\end{equation*}
where $\bm{\kappa}_j$ and $\sigma$ denote $\kappa(\bm{x}_j , H_\times)$ and $\sigma_\times$, respectively. Let $\mathbf{Z} = [ \bm{z}_1^\top ; \bm{z}_2^\top ; \dots ; \bm{z}_{H_\times}^\top ]$ and $\bm{\alpha} = ( \alpha_1 ; \alpha_2 ; \dots ; \alpha_{H_\times} )$. We prove the theorem by discussion. \\ 
\textbf{Case 1.} There exists $k_0 \in [H_\times]$, such that $\alpha_{k_0} \neq 0$. Since the loss $\hat{L}(\mathbf{Z},\bm{\alpha})$ is positive and $l(0) = 0$, there exists $j_0 \in [n]$, such that $\bm{\alpha}^\top \mathrm{Re} [ \sigma ( \mathbf{Z} \bm{\kappa}_{j_0} ) ] \neq y_{j_0}$. Without loss of generality, we only consider the case of $\bm{\alpha}^\top \mathrm{Re} [ \sigma ( \mathbf{Z} \bm{\kappa}_{j_0} ) ] > y_{j_0}$ in this proof. The other case can be proven similarly. In view of the condition of all samples being linearly independent and the definition of $\kappa$ in Eq.~\eqref{eq: definition of phi}, one knows that $\{ \bm{\kappa}_j \}_{j=1}^n$ are linearly independent. Thus, there exists a non-zero vector $\bm{v} \in \mathbb{C}^{H_\times}$, such that $\bm{v}^\top \bm{\kappa}_{j_0} \neq 0$ and $\bm{v}^\top \bm{\kappa}_j = 0$ hold for any $j \in [n] \backslash \{j_0\}$. Let $\Delta \bm{\alpha} = \bm{0}$, $\bm{z}_k = \bm{0}$ for any $k \in [H_\times] \backslash \{k_0\}$, and $\bm{z}_{k_0} = c \bm{v}$ where $c \in \mathbb{C}$ is a complex-valued variable. Then the change in loss becomes a function of $c$ as follows 
\begin{equation*}
    \begin{aligned}
        \Delta \hat{L} \left( c \right) 
        =&~ l \Big( \bm{\alpha}^\top \mathrm{Re} \left[ \sigma \left( \mathbf{Z} \bm{\kappa}_{j_0} \right) \right] + \alpha_{k_0} \mathrm{Re} \left[ \sigma \left( \left( \bm{z}_{k_0} + c \bm{v} \right)^\top \bm{\kappa}_{j_0} \right) \right] \\ 
        &- \alpha_{k_0} \mathrm{Re} \left[ \sigma \left( \bm{z}_{k_0}^\top \bm{\kappa}_{j_0} \right) \right] - y_{j_0} \Big) - l \left( \bm{\alpha}^\top \mathrm{Re} \left[ \sigma \left( \mathbf{Z} \bm{\kappa}_{j_0} \right) \right] - y_{j_0} \right) , 
    \end{aligned}
\end{equation*}
where the equality holds since the output of $\mathrm{FTNet}$ on $\bm{x}^{(j)}$ remains the same for any $j \neq j_0$. Since $\alpha_{k_0} \neq 0$, $\bm{v}^\top \bm{\kappa}_{j_0} \neq 0$, and $\sigma$ is holomorphic and not constant, one knows that $\alpha_{k_0} \sigma ( ( \bm{z}_{k_0} + c \bm{v} )^\top \bm{\kappa}_{j_0} )$ is holomorphic and not constant w.r.t. $c$. Then Lemma~\ref{lem: neighborhood of holomorphic functions} implies that there exists $c \leqslant \delta / \Vert \bm{v} \Vert_2$, s.t.
\begin{equation*}
    \mathrm{Re} \left[ \alpha_{k_0} \sigma \left( \left( \bm{z}_{k_0} + c \bm{v} \right)^\top \bm{\kappa}_{j_0} \right) \right] 
    < \mathrm{Re} \left[ \alpha_{k_0} \sigma \left( \bm{z}_{k_0}^\top \bm{\kappa}_j \right) \right]
\end{equation*}
and
\begin{equation}
    \label{eq: relation of re, proof of no local minima}
    \mathrm{Re} \left[ \alpha_{k_0} \sigma \left( \left( \bm{z}_{k_0} + c \bm{v} \right)^\top \bm{\kappa}_{j_0} \right) \right]
    \geqslant \mathrm{Re} \left[ \alpha_{k_0} \sigma \left( \bm{z}_{k_0}^\top \bm{\kappa}_j \right) \right] - \bm{\alpha}^\top \mathrm{Re} \left[ \sigma \left( \mathbf{Z} \bm{\kappa}_{j_0} \right) \right] + y_{j_0} , 
\end{equation}
where Eq.~\eqref{eq: relation of re, proof of no local minima} can be satisfied based on the continuity of holomorphic functions. Thus, one has $\Vert \Delta \mathbf{Z} \Vert_F + \Vert \Delta \bm{\alpha} \Vert_2 = \Vert c \bm{v} \Vert_2 = |c| \Vert \bm{v} \Vert_2 \leqslant \delta$ and $\Delta \hat{L} ( c ) < 0$ since the loss function $l$ is strictly increasing on $(0,+\infty)$. \\ 
\textbf{Case 2.} For any $k \in [H_\times]$, $\alpha_k = 0$. Let $\Delta \bm{z}_k = \bm{0}$ and $\Delta \alpha_k = 0$ for any $k \in [H_\times] \backslash \{1\}$. Then the change in loss becomes a function of $\Delta \bm{z}_1$ and $\Delta \alpha_1$ as follows 
\begin{equation*}
    \Delta \hat{L} 
    = \sum_{j=1}^n l \left( \Delta \alpha_1 \mathrm{Re} \left[ \sigma \left( \left( \bm{z}_1 + \Delta \bm{z}_1 \right)^\top \bm{\kappa}_j \right) \right] - y_j \right) - l \left( - y_j \right) . 
\end{equation*}
The proof of this case is divided into several steps. \\ 
\textbf{Step 2.1.} We rewrite the change of loss in a power series form. Since the loss function $l$ is analytic, there exist coefficients $\{ c_p \}_{p=0}^\infty$, such that $l(y) = \sum_{p=0}^\infty c_p y^p$ holds for any $y \in \mathbb{R}$. Then the change of loss can be rewritten as 
\begin{equation}
    \label{eq: change of loss, in proof of no local minimum, step 2.1}
    \begin{aligned}
        \Delta \hat{L} 
        =&~ \sum_{j=1}^n \sum_{p=1}^\infty \sum_{q=1}^p c_p \binom{p}{q} \left( - y_j \right)^{p-q} \left( \Delta \alpha_1 \mathrm{Re} \left[ \sigma \left( \left( \bm{z}_1 + \Delta \bm{z}_1 \right)^\top \bm{\kappa}_j \right) \right] \right)^q \\ 
        =&~ \sum_{q=1}^\infty \sum_{j=1}^n \sum_{p=q}^\infty c_p \binom{p}{q} \left( - y_j \right)^{p-q} \left( \Delta \alpha_1 \right)^q \left( \mathrm{Re} \left[ \sigma \left( \left( \bm{z}_1 + \Delta \bm{z}_1 \right)^\top \bm{\kappa}_j \right) \right] \right)^q \\ 
        =&~ \sum_{q=1}^\infty C_q \left( \Delta \alpha_1 \right)^q , 
    \end{aligned}
\end{equation}
where the first equality holds from the binomial expansion, the second equality holds by changing the order of summation, and $C_q$ is a function of $\Delta \bm{z}_1$ defined by 
\begin{equation}
    \label{eq: def of cq, in proof of no local minimum, step 2.1}
    C_q 
    = \sum_{j=1}^n \sum_{p=q}^\infty c_p \binom{p}{q} R^q \left( - y_j \right)^{p-q}
\end{equation}
with $R = \mathrm{Re} [ \sigma ( \left( \bm{z}_1 + \Delta \bm{z}_1 \right)^\top \bm{\kappa}_j ) ]$. \\
\textbf{Step 2.2.} We prove that $C_1$ defined in Eq.~\eqref{eq: def of cq, in proof of no local minimum, step 2.1} is not always zero. For $q=1$, it is observed that 
\begin{equation*}
    C_1 
    = \sum_{j=1}^n \sum_{p=1}^\infty c_p p \mathrm{Re} \left[ \sigma \left( \left( \bm{z}_1 + \Delta \bm{z}_1 \right)^\top \bm{\kappa}_j \right) \right] \left( - y_j \right)^{p-1}
    = \mathrm{Re} \left[ \sum_{j=1}^n l'(-y_j) \sigma \left( \left( \bm{z}_1 + \Delta \bm{z}_1 \right)^\top \bm{\kappa}_j \right) \right] . 
\end{equation*}
Since the loss function $l$ is a well-posed regression loss function, one knows that the equation $l'(y) = 0$ has a unique solution $y=0$. Since $\bm{\alpha} = 0$, all outputs of $\mathrm{F}\text{-}\mathrm{FTNet}$ are $0$. In view of positive loss, one knows that $\{ y_j \}_{j=1}^n$ are not all $0$, which indicates that $\{ l'(-y_j) \}_{j=1}^n$ are not all $0$. Since $\{ \bm{\kappa}_j \}_{j=1}^n$ are linearly independent, they are different. Note that $\sigma$ is holomorphic and not polynomial, Lemma~\ref{lem: not a constant} indicates that $\sum_{j=1}^n l'(-y_j) \sigma ( ( \bm{z}_1 + \Delta \bm{z}_1 )^\top \bm{\kappa}_j )$ is not a constant. Thus, there exists $\Delta \bm{z}_1$, such that $\Vert \Delta \bm{z}_1 \Vert_2 \leqslant \delta / 2$ and $C_1 \neq 0$. \\ 
\textbf{Step 2.3.} We give upper bounds for $\{ C_q \}_{q=2}^\infty$. Provided $\Delta \bm{z}_1$ in Step 2.2., we define $a = \max_{j \in [n]} | \mathrm{Re} [ \sigma ( ( \bm{z}_1 + \Delta \bm{z}_1 )^\top \bm{\kappa}_j ) ] |$. Let $b = \max_{j \in [n]} |y_j|$ for labels $\{ y_j \}_{j=1}^\infty$. Since the loss function $l$ is analytic on $\mathbb{R}$, the convergence radius of its Taylor series should be infinity. Thus, one has $\lim \sup_{p \rightarrow \infty} |c_p|^{1/p} = 0$ from the Cauchy-Hadamard theorem. Furthermore, there exists $d>0$, s.t. $|c_p| \leqslant d / (4b)^p$ holds for any $p \geqslant 2$. Using these notations, coefficients $\{ C_q \}_{q=2}^\infty$ can be bounded by 
\begin{equation}
    \label{eq: upper bound of cq, in proof of no local minimum}
    |C_q| 
    \leqslant \sum_{j=1}^n \sum_{p=q}^\infty |c_p| \binom{p}{q} \left| \mathrm{Re} \left[ \sigma \left( \left( \bm{z}_1 + \Delta \bm{z}_1 \right)^\top \bm{\kappa}_j \right) \right] \right|^q \left| y_j \right|^{p-q} \\ 
    \leqslant \sum_{p=q}^\infty \frac{nd}{4^p} \binom{p}{q} \left( \frac{a}{b} \right)^q , 
\end{equation}
where the first inequality holds from the triangle inequality. \\
\textbf{Step 2.4.} We choose a proper $\Delta \alpha_1$ and give an upper bound for $\Delta \hat{L}$. Let $\Delta \alpha_1 = - \mathrm{sign}(C_1) b/(ka)$, where $k \geqslant 1$ is a coefficient determined later. Thus, the change of loss in Eq.~\eqref{eq: change of loss, in proof of no local minimum, step 2.1} can be rewritten as 
\begin{equation*}
    \Delta \widehat{L} 
    \leqslant C_1 \Delta \alpha_1 + \sum_{q=2}^\infty |C_q| \left| \Delta \alpha_1 \right|^q
    \leqslant - \frac{|C_1|b}{ka} + \sum_{p=2}^\infty \sum_{q=2}^p \frac{nd}{4^p} \binom{p}{q} \frac{1}{k^q}
    \leqslant - \frac{|C_1|b}{ka} + \frac{nd}{2k^2} , 
\end{equation*}
where the first inequality holds according to the triangle inequality, the second inequality holds based on Eq.~\eqref{eq: upper bound of cq, in proof of no local minimum}, the choice of $\Delta \alpha_1$, and changing the order of summation, and the third inequality holds because of $k \geqslant 1$. We employ
\begin{equation*}
    k = \max \left\{ 1 , \frac{nda}{|C_1|b}, \frac{2b}{a \delta} \right\} , 
\end{equation*}
and thus, one has $\Delta \hat{L} < 0$ and 
\begin{equation*}
    \left\Vert \Delta \mathbf{Z} \right\Vert_F + \left\Vert \Delta \bm{\alpha} \right\Vert_2 = \left\Vert \Delta \bm{z}_1 \right\Vert_2 + \left| \Delta \alpha_1 \right| \leqslant \delta / 2 + \delta / 2 = \delta . 
\end{equation*}
Combining the results in all cases completes the proof. $\hfill \square$ 

\section{Conclusion and Prospect}
\label{sec: conclusion and prospect}

This work investigates the theoretical properties of $\mathrm{FTNet}$ via approximation and local minima. The main conclusions are three folds. Firstly, we prove the universal approximation of $\mathrm{F}\text{-}\mathrm{FTNet}$ and $\mathrm{R}\text{-}\mathrm{FTNet}$, which guarantees the possibility of expressing any continuous function and any discrete-time open dynamical system on any compact set arbitrarily well, respectively. Secondly, we claim the approximation-complexity advantages and worst-case guarantees of one-hidden-layer $\mathrm{F}\text{-}\mathrm{FTNet}$/$\mathrm{R}\text{-}\mathrm{FTNet}$ over $\mathrm{FNN}$/$\mathrm{RNN}$, i.e., $\mathrm{F}\text{-}\mathrm{FTNet}$ and $\mathrm{R}\text{-}\mathrm{FTNet}$ can express some functions with an exponentially fewer number of hidden neurons and can express a function with the same order of hidden neurons in the worst case, compared with $\mathrm{FNN}$ and $\mathrm{RNN}$, respectively. Thirdly, we provide the feasibility of optimizing $\mathrm{F}\text{-}\mathrm{FTNet}$ to the global minimum using local search algorithms, i.e., the loss surface of one-hidden-layer $\mathrm{F}\text{-}\mathrm{FTNet}$ has no suboptimal local minimum using general activations and loss functions. Our theoretical results take one step towards the theoretical understanding of $\mathrm{FTNet}$, which exhibits the possibility of ameliorating $\mathrm{FTNet}$. In the future, it is important to investigate other properties or advantages of $\mathrm{FTNet}$ beyond classical neural networks, such as from the perspectives of optimization and generalization.

\section*{Acknowledgements}

This research was supported by the National Key Research and Development Program of China (2020AAA0109401), NSFC (61921006, 62176117), and Collaborative Innovation Center of Novel Software Technology and Industrialization. The authors would like to thank Shoucheng Yu for helpful discussions about complex analysis and Shen-Huan Lyu, Peng Tan, and Zhi-Hao Tan for feedback on drafts of the paper.

\appendix

\section{Complete Proof of Eq.~\eqref{eq: ua of addftnet, main body}}

\noindent \textbf{Proof.} We only demonstrate the proof when $\sigma_1$ and $\sigma_2$ are continuous for simplicity. The case of almost everywhere continuous activation functions can be proven with a slight modification. The proof is divided into several steps. \\ 
\textbf{Step 1}. We prove that $\mathrm{FNN}$ with activation functions $\sigma_1$ and $\sigma_2$ can approximate the state transition function $\varphi$ of $\mathrm{DODS}$, defined in Eq.~\eqref{eq: mapping of dods}. Since the hidden state transition function $\varphi$ is continuous, and the image of a continuous function defined on the compact set $K$ is a compact set, there exists a convex compact set $K_1 \in \mathbb{R}^{H_\mathrm{D}}$, such that $\bm{h}_t \in K_1$ holds for any $t \in \{ 0,1,\cdots,T \}$. Let $B_\infty(A,r) = \mathop{\cup}_{a \in A} \{ b \mid \Vert b - a \Vert_\infty \leqslant r \}$ denote the neighborhood of the set $A$ with radius $r$, and $K_2 = K \times B_\infty(K_1,1)$ is the Cartesian product of $K$ and $B_\infty(K_1,1)$. It is easy to check that $K_2$ is convex and compact. Let $\bm{x} \in \mathbb{R}^I$ and $\bm{h} \in \mathbb{R}^{H_\mathrm{D}}$. Since both $\sigma_1$ and $\sigma_2$ are continuous almost everywhere and not polynomial almost everywhere, Lemma~\ref{lem: ua with row independent matrix} indicates that for any $\varepsilon_1 > 0 $, there exist $ H_1 , H_2 \in \mathbb{N}^+ ,  \mathbf{A}_1 \in \mathbb{R}^{H_1 \times I} , \mathbf{B}_1 \in \mathbb{R}^{H_1 \times H_\mathrm{D}} , \mathbf{C}_1 \in \mathbb{R}^{H_\mathrm{D} \times H_1} , \bm{\theta}_1 \in \mathbb{R}^{H_1} , \mathbf{A}_2 \in \mathbb{R}^{H_2 \times I} , \mathbf{B}_2 \in \mathbb{R}^{H_2 \times H_\mathrm{D}} , \mathbf{C}_2 \in \mathbb{R}^{H_\mathrm{D} \times H_2}$, and $\bm{\theta}_2 \in \mathbb{R}^{H_2} $, where $\mathbf{C}_1$ and $\mathbf{C}_2$ are row independent, such that 
\begin{equation}
    \label{ua of phi}
    \begin{gathered}
        \sup_{(\bm{x},\bm{h}) \in K_2} \Vert \varphi(\bm{x},\bm{h}) - \mathbf{C}_1 \sigma_1 \left( \mathbf{A}_1 \bm{x} + \mathbf{B}_1 \bm{h} - \bm{\theta}_1 \right) \Vert_\infty \leqslant \varepsilon_1 , \\ 
        \sup_{(\bm{x},\bm{h}) \in K_2} \Vert \varphi(\bm{x},\bm{h}) - \mathbf{C}_2 \sigma_2 \left( \mathbf{A}_2 \bm{x} + \mathbf{B}_2 \bm{h} - \bm{\theta}_1 \right) \Vert_\infty \leqslant \varepsilon_1 . 
    \end{gathered}
\end{equation}
\textbf{Step 2}. We prove that $\mathrm{RNN}$ using the same weight matrices as $\mathrm{FNN}$ in Eq.~\eqref{ua of phi} can approximate $\bm{h}_t$, the hidden state of $\mathrm{DODS}$. Let $\bm{p}_0^{(1)} = \bm{q}_0^{(1)} = \bm{h}_0$. For any $t \in [T]$, define  
\begin{equation}
    \label{eq: rnn in step 2, in ua of addftnet}
    \begin{gathered}
        \bm{p}_t^{(1)} = \mathbf{C}_1 \sigma_1 \left( \mathbf{A}_1 \bm{x}_t + \mathbf{B}_1 \bm{p}_{t-1}^{(1)} - \bm{\theta}_1 \right) \in \mathbb{R}^{H_\mathrm{D}} , \\ 
        \bm{q}_t^{(1)} = \mathbf{C}_2 \sigma_2 \left( \mathbf{A}_2 \bm{x}_t + \mathbf{B}_2 \bm{q}_{t-1}^{(1)} - \bm{\theta}_2 \right) \in \mathbb{R}^{H_\mathrm{D}} . 
    \end{gathered}
\end{equation}
The above $\bm{p}_t^{(1)}$ and $\bm{q}_t^{(1)}$ are outputs of two different $\mathrm{RNN}$s. We then prove that $\bm{p}_t^{(1)}$ and $\bm{q}_t^{(1)}$ can approximate $\bm{h}_t$. Let $ u : [0,+\infty) \rightarrow \mathbb{R} $ be defined as 
\begin{equation*}
    u(a) = \sup \left\{ \Vert \varphi(\bm{y}) - \varphi(\bm{z}) \Vert_\infty \mid \bm{y} , \bm{z} \in K_2 , \Vert \bm{y} - \bm{z} \Vert_\infty \leqslant a \right\} . 
\end{equation*}
From Lemma~\ref{lem: u is continuous}, $u(a)$ is continuous. For any $ t $, if $ \Vert \bm{h}_{t-1} - \bm{p}_{t-1}^{(1)} \Vert_\infty \leqslant 1 $, then $ ( \bm{x}_t , \bm{p}_{t-1}^{(1)} ) \in K_2 $, and one has 
\begin{equation*}
    \begin{aligned}
        \left\Vert \bm{h}_t - \bm{p}_t^{(1)} \right\Vert_\infty
        &= \left\Vert \varphi(\bm{x}_t , \bm{h}_{t-1}) - \mathbf{C}_1 \sigma_1 \left( \mathbf{A}_1 \bm{x}_t + \mathbf{B}_1 \bm{p}_{t-1}^{(1)} - \bm{\theta}_1 \right) \right\Vert_\infty \\ 
        &\leqslant \left\Vert \varphi(\bm{x}_t , \bm{h}_{t-1}) - \varphi \left( \bm{x}_1 , \bm{p}_{t-1}^{(1)} \right) \right\Vert_\infty \\
        & \quad + \left\Vert \varphi \left(\bm{x}_1 , \bm{p}_{t-1}^{(1)} \right) - \mathbf{C}_1 \sigma_1 \left( \mathbf{A}_1 \bm{x}_t + \mathbf{B}_1 \bm{p}_{t-1}^{(1)} - \bm{\theta}_1 \right) \right\Vert_\infty \\ 
        &\leqslant u \left( \left\Vert \bm{h}_{t-1} - \bm{p}_{t-1}^{(1)} \right\Vert_\infty \right) + \varepsilon_1 , 
    \end{aligned}
\end{equation*}
where the first equality holds from the definitions of $\mathrm{DODS}$ and $\bm{p}_t^{(1)}$, the first inequality holds because of the triangle inequality, and the second inequality holds based on the definition of $u(a)$, $ ( \bm{x}_t , \bm{p}_{t-1}^{(1)} ) \in K_2 $, and Eq.~\eqref{ua of phi}. Let $ a_0 = 0 $. For any $ t \in [T] $, we define $a_t = u(a_{t-1}) + \varepsilon_1$. Then Lemma~\ref{lem: limit is 0} indicates $ \lim_{\varepsilon_1 \rightarrow 0_+} a_t = 0 $ for any $t \in [T]$, i.e., for any $\varepsilon_2 \in (0,1)$, there exists $\delta_1(\varepsilon_2) > 0$, such that for any $\varepsilon_1 \leqslant \delta_1(\varepsilon_2)$, $ a_t \leqslant \varepsilon_2 $ holds for any $t \in [T]$. When $ \varepsilon_1 \leqslant \delta_1(\varepsilon_2) $, it is easy to see that $\Vert \bm{h}_t - \bm{p}_t^{(1)} \Vert_\infty \leqslant a_t \leqslant \varepsilon_2$ holds for any $t \in [T]$. The same conclusion can be proven for $\bm{q}_t^{(1)}$ in the same way. Thus, for any $ \varepsilon_1 \leqslant \delta_1(\varepsilon_2) $, one has 
\begin{equation}
    \label{relation between m and p1 q1}
    \max_{t \in [T]} \left\Vert \bm{h}_t - \bm{p}_t^{(1)} \right\Vert_\infty \leqslant \varepsilon_2 
    \quad 
    \text{and} 
    \quad 
    \max_{t \in [T]} \left\Vert \bm{h}_t - \bm{q}_t^{(1)} \right\Vert_\infty \leqslant \varepsilon_2 . 
\end{equation}
\textbf{Step 3}. Transformation is used to eliminate the matrices $\mathbf{C}_1$ and $\mathbf{C}_2$ in Eq.~\eqref{eq: rnn in step 2, in ua of addftnet}, which is the preparation to approximate $\bm{h}_t$ using additive $\mathrm{FTNet}$. Since $\mathbf{C}_1 , \mathbf{C}_2 $ are row independent, both $ \mathbf{C}_1 \bm{x} = \bm{p}_0^{(1)} $ and $ \mathbf{C}_2 \bm{x} = \bm{q}_0^{(1)} $ have solutions. Let $ \bm{p}_0^{(2)}$ and $\bm{q}_0^{(2)} $ be the solutions of the above equations, respectively, i.e., $\mathbf{C}_1 \bm{p}_0^{(2)} = \bm{p}_0^{(1)}$ and $\mathbf{C}_2 \bm{q}_0^{(2)} = \bm{q}_0^{(1)}$. Define 
\begin{equation}
    \label{eq: rnn in step 3, in ua of addftnet}
    \begin{gathered}
        \bm{p}_t^{(2)} = \sigma_1 \left( \mathbf{A}_1 \bm{x}_t + \mathbf{B}_1 \mathbf{C}_1 \bm{p}_{t-1}^{(2)} - \bm{\theta}_1 \right) \in \mathbb{R}^{H_1} , \\ 
        \bm{q}_t^{(2)} = \sigma_2 \left( \mathbf{A}_2 \bm{x}_t + \mathbf{B}_2 \mathbf{C}_2 \bm{q}_{t-1}^{(2)} - \bm{\theta}_2 \right) \in \mathbb{R}^{H_2} . 
    \end{gathered}
\end{equation}
We claim that, for any $t \in \{0,1,\cdots,T\} $, 
\begin{equation}
    \label{eq: p1 = c1 p2 , q1 = c2 q2}
    \bm{p}_t^{(1)} = \mathbf{C}_1 \bm{p}_t^{(2)} , \quad \bm{q}_t^{(1)} = \mathbf{C}_2 \bm{q}_t^{(2)} . 
\end{equation}
Since the proof of $\bm{q}_t^{(2)}$ is similar to that of $\bm{p}_t^{(2)}$, we only give the proof of $\bm{p}_t^{(2)}$ using mathematical induction as follows. 
\begin{enumerate}
    \item For $t=0$, the claim holds from the definition of $ \bm{p}_0^{(2)} $. 
    \item Suppose that the claim holds for $t=k$, where $k \in \{0,1,\cdots,T-1\}$. Thus, one has 
    \begin{equation*}
        \begin{aligned}
            \bm{p}_{k+1}^{(1)} 
            =&~ \mathbf{C}_1 \sigma_1 \left( \mathbf{A}_1 \bm{x}_{k+1} + \mathbf{B}_1 \bm{p}_k^{(1)} - \bm{\theta}_1 \right) \\ 
            =&~ \mathbf{C}_1 \sigma_1 \left( \mathbf{A}_1 \bm{x}_{k+1} + \mathbf{B}_1 \mathbf{C}_1 \bm{p}_k^{(2)} - \bm{\theta}_1 \right) \\ 
            =&~ \mathbf{C}_1 \bm{p}_{k+1}^{(2)} , 
        \end{aligned}
    \end{equation*}
    where the first equality holds from the definition of $\bm{p}_t^{(1)}$ with $t=k+1$ in Eq.~\eqref{eq: rnn in step 2, in ua of addftnet}, the second equality holds because of the induction hypothesis, and the third equality holds based on the definition of $\bm{p}_t^{(2)}$ with $t=k+1$. Thus, the claim holds for $t=k+1$. 
\end{enumerate}
\textbf{Step 4}. We prove that additive $\mathrm{FTNet}$ can approximate $\bm{h}_t$ by unifying the weight matrices in Eq.~\eqref{eq: rnn in step 3, in ua of addftnet}. Let $ H_3 = H_1 + H_2 $. For any $ t \in [T] $, define 
\begin{equation}
    \label{eq: addftnet in step 4, in ua of addftnet}
    \begin{gathered}
        \bm{p}_t^{(3)} = \sigma_1 \left( \mathbf{A}_3 \bm{x}_t + \mathbf{B}_3 \bm{q}_{t-1}^{(3)} - \bm{\theta}_3 \right) \in \mathbb{R}^{H_3} , \\ 
        \bm{q}_t^{(3)} = \sigma_2 \left( \mathbf{A}_3 \bm{x}_t + \mathbf{B}_3 \bm{q}_{t-1}^{(3)} - \bm{\theta}_3 \right) \in \mathbb{R}^{H_3} , 
    \end{gathered}
\end{equation}
where 
\begin{equation*}
    \bm{q}_0^{(3)} = \left[ 
        \begin{array}{c}
            \bm{0} \\ 
            \bm{q}_0^{(2)} 
        \end{array}
    \right] , 
    \quad 
    \mathbf{A}_3 = \left[ 
        \begin{array}{c}
            \mathbf{A}_1 \\ 
            \mathbf{A}_2 
        \end{array}
    \right] ,
    \quad
    \mathbf{B}_3 = \left[ 
        \begin{array}{cc}
            \bm{0} & \mathbf{B}_1 \mathbf{C}_2 \\ 
            \bm{0} & \mathbf{B}_2 \mathbf{C}_2 
        \end{array}
    \right] , 
    \quad 
    \bm{\theta}_3 = \left[ 
        \begin{array}{c}
            \bm{\theta}_1 \\ 
            \bm{\theta}_2 
        \end{array}
    \right] . 
\end{equation*}
For $\bm{q}_t^{(3)}$, we claim that $\bm{q}_t^{(3)} = [ \bm{\times}_t ; \bm{q}_t^{(2)} ]$ holds for any $t \in [T]$, where $\bm{\times}_t \in \mathbb{R}^{H_1 \times 1}$ is a vector that we do not care, because it has no contribution to the iteration or output in the above additive $\mathrm{FTNet}$. We prove the claim about $\bm{q}_t^{(3)}$ by mathematical induction as follows. 
\begin{enumerate}
    \item For $t=1$, one has 
    \begin{equation*}
        \begin{aligned}
            \bm{q}_1^{(3)} 
            =& \sigma_2 \left( \mathbf{A}_3 \bm{x}_1 + \mathbf{B}_3 \bm{q}_0^{(3)} - \bm{\theta}_3 \right) \\ 
            =& \sigma_2 \left( \left[ 
                \begin{array}{c}
                    \mathbf{A}_1 \bm{x}_1 + \mathbf{B}_1 \mathbf{C}_2 \bm{q}_0^{(2)} - \bm{\theta}_1 \\ 
                    \mathbf{A}_2 \bm{x}_1 + \mathbf{B}_2 \mathbf{C}_2 \bm{q}_0^{(2)} - \bm{\theta}_2 
                \end{array}
            \right] \right) \\ 
            =& \left[ \times_1 ; \bm{q}_1^{(2)} \right] , 
        \end{aligned}
    \end{equation*}
    where the first equality holds according to the definition of $\bm{q}_t^{(3)}$ with $t=1$ in Eq.~\eqref{eq: addftnet in step 4, in ua of addftnet}, the second equality holds based on the definitions of $\mathbf{A}_3 , \mathbf{B}_3 , \bm{\theta}_3 $, and the third equality holds from the definition of $\bm{q}_t^{(2)}$ with $t=1$ in Eq.~\eqref{eq: rnn in step 3, in ua of addftnet}. Thus, the claim holds for $t=1$. 
    \item Suppose that the claim holds for $t=k$ where $k \in [T-1]$. Thus, one has 
    \begin{equation*}
        \begin{aligned}
            \bm{q}_{k+1}^{(3)} 
            =& \sigma_2 \left( \mathbf{A}_3 \bm{x}_{k+1} + \mathbf{B}_3 \bm{q}_k^{(3)} - \bm{\theta}_3 \right) \\ 
            =& \sigma_2 \left( \left[ 
                \begin{array}{c}
                    \mathbf{A}_1 \bm{x}_{k+1} + \mathbf{B}_1 \mathbf{C}_2 \bm{q}_k^{(2)} - \bm{\theta}_1 \\ 
                    \mathbf{A}_2 \bm{x}_{k+1} + \mathbf{B}_2 \mathbf{C}_2 \bm{q}_k^{(2)} - \bm{\theta}_2 
                \end{array}
            \right] \right) \\ 
            =& \left[ \times_{k+1} ; \bm{q}_{k+1}^{(2)} \right] , 
        \end{aligned}
    \end{equation*}
    where the first equality holds from the definition of $\bm{q}_t^{(3)}$ with $t=k+1$, the second equality holds because of the definitions of $\mathbf{A}_3 , \mathbf{B}_3 , \bm{\theta}_3 $, and the third equality holds based on the definition of $\bm{q}_t^{(2)}$ with $t=k+1$. Thus, the claim holds for $t=k+1$. 
\end{enumerate}
We then study the property of $\bm{p}_t^{(3)}$. Let 
\begin{equation*}
    \mathbf{C}_3 = \left[ 
        \begin{array}{cc}
            \mathbf{C}_1 & \bm{0}^{H_\mathrm{D} \times H_2} 
        \end{array}
    \right] . 
\end{equation*}
If $ \Vert \bm{h}_{t-1} - \bm{q}_{t-1}^{(1)} \Vert_\infty \leqslant 1 $, then $ ( \bm{x}_t , \bm{q}_{t-1}^{(1)} ) \in K_2 $, and one has 
\begin{equation*}
    \begin{aligned}
        \left\Vert \bm{h}_t - \mathbf{C}_3 \bm{p}_t^{(3)} \right\Vert_\infty
        &= \left\Vert \varphi \left( \bm{x}_t , \bm{h}_{t-1} \right) - \mathbf{C}_1 \sigma_1 \left( \mathbf{A}_1 \bm{x}_t + \mathbf{B}_1 \mathbf{C}_2 \bm{q}_{t-1}^{(2)} - \bm{\theta}_1 \right) \right\Vert_\infty \\ 
        &\leqslant \left\Vert \varphi \left( \bm{x}_t , \bm{h}_{t-1} \right) - \varphi \left( \bm{x}_t , \bm{q}_{t-1}^{(1)} \right) \right\Vert_\infty \\ 
        & \quad + \left\Vert \varphi \left( \bm{x}_t , \bm{q}_{t-1}^{(1)} \right) - \mathbf{C}_1 \sigma_1 \left( \mathbf{A}_1 \bm{x}_t + \mathbf{B}_1 \bm{q}_{t-1}^{(1)} - \bm{\theta}_1 \right) \right\Vert_\infty \\ 
        &\leqslant u \left( \varepsilon_2 \right) + \varepsilon_1 , 
    \end{aligned}
\end{equation*}
where the first equality holds because of the definitions of $\mathrm{DODS}$, $\mathbf{C}_3$, and $\bm{p}_t^{(3)}$, the first inequality holds based on the triangle inequality and Eq.~\eqref{eq: p1 = c1 p2 , q1 = c2 q2}, and the second inequality holds based on the definition of $u(a)$, $( \bm{x}_t , \bm{q}_{t-1}^{(1)} ) \in K_2$, Eqs.~\eqref{ua of phi} and~\eqref{relation between m and p1 q1}. Then according to the continuity of $u(a)$ and $u(0)=0$, for any $\varepsilon_3 > 0$, there exist $\delta_2(\varepsilon_3)$ and $\delta_3(\varepsilon_3)$, such that if $\varepsilon_1 \leqslant \delta_2(\varepsilon_3)$ and $\varepsilon_2 \leqslant \delta_3(\varepsilon_3)$, one has 
\begin{equation}
    \label{eq: relation between m and c3 p3}
    \max_{t \in [T]} \left\Vert \bm{h}_t - \mathbf{C}_3 \bm{p}_t^{(3)} \right\Vert_\infty \leqslant \varepsilon_3 . 
\end{equation}
\textbf{Step 5}. The output function $\psi$ in Eq.~\eqref{eq: mapping of dods} can be approximated by an $\mathrm{FNN}$. Let $\tau_3$ be any continuous non-polynomial function. According to Lemma~\ref{lem: ua of fnn}, for any $\varepsilon_4 > 0$, there exist $H_4 \in \mathbb{N}^+ , \mathbf{A}_4 \in \mathbb{R}^{H_4 \times H_\mathrm{D}} , \mathbf{B}_4 \in \mathbb{R}^{O \times H_4} , \bm{\theta}_4 \in \mathbb{R}^{H_4 \times 1} $, s.t. 
\begin{equation}
    \label{eq: ua of psi}
    \sup_{\bm{h} \in B(K_1,1)} \left\Vert \psi(\bm{h}) - \mathbf{B}_4 \tau_3 \left( \mathbf{A}_4 \bm{h} - \bm{\theta}_4 \right) \right\Vert_\infty \leqslant \varepsilon_4 . 
\end{equation}
For any $t \in [T]$, define $\bm{y}_t^{(1)} = \mathbf{B}_4 \tau_3 ( \mathbf{A}_4 \mathbf{C}_3 \bm{p}_t^{(3)} - \bm{\theta}_4 )$. Substituting the definition of $\bm{p}_t^{(3)}$ in Eq.~\eqref{eq: addftnet in step 4, in ua of addftnet} into the above definition, one has, for any $t \in [T]$,
\begin{equation}
    \label{eq: temp 1, proof of ua of r-ftnet, firstly step 5}
    \bm{y}_t^{(1)} = \mathbf{B}_4 \tau_3 \left( \mathbf{A}_4 \mathbf{C}_1 \sigma_1 \left( \mathbf{A}_1 \bm{x}_t + \mathbf{B}_1 \mathbf{C}_2 \bm{q}_{t-1}^{(2)} - \bm{\theta}_3 \right) - \bm{\theta}_4 \right) . 
\end{equation}
Since $\sigma_2$ is continuous, and $\bm{x}_t \in K$ holds for any $t \in [T]$, there exists a compact set $K_3$, s.t. $\bm{q}_t^{(2)} \in K_3$ holds for any $t \in [T]$. Since $\sigma_1$ is continuous and not polynomial, Lemma~\ref{lem: ua of fnn} implies that for any $\varepsilon_5 > 0 $, there exist $H_5 \in \mathbb{N}^+ , \mathbf{A}_5 \in \mathbb{R}^{H_5 \times I} , \mathbf{B}_5 \in \mathbb{R}^{H_5 \times H_2} , \mathbf{C}_5 \in \mathbb{R}^{O \times H_5}$, and $\bm{\theta}_5 \in \mathbb{R}^{H_5} $, s.t. 
\begin{equation}
    \label{eq: temp 2, proof of ua of r-ftnet, firstly step 5}
    \sup_{(\bm{x},\bm{q}) \in K \times K_3} \big\Vert \mathbf{B}_4 \tau_3 \left( \mathbf{A}_4 \mathbf{C}_1 \sigma_1 \left( \mathbf{A}_1 \bm{x} + \mathbf{B}_1 \mathbf{C}_2 \bm{q} - \bm{\theta}_3 \right) - \bm{\theta}_4 \right) - \mathbf{C}_5 \sigma_1 \left( \mathbf{A}_5 \bm{x} + \mathbf{B}_5 \bm{q} - \bm{\theta}_5 \right) \big\Vert_\infty \leqslant \varepsilon_5 .
\end{equation}
For any $t \in [T]$, define 
\begin{equation}
    \label{eq: def of pt5, proof of ua of r-ftnet, firstly step 5}
    \begin{gathered}
        \bm{p}_t^{(5)} = \sigma_1 \left( \mathbf{A}_5 \bm{x}_t + \mathbf{B}_5 \bm{q}_{t-1}^{(2)} - \bm{\theta}_5 \right) \in \mathbb{R}^{H_5} , \\ 
        \bm{q}_t^{(5)} = \sigma_2 \left( \mathbf{A}_5 \bm{x}_t + \mathbf{B}_5 \bm{q}_{t-1}^{(2)} - \bm{\theta}_5 \right) \in \mathbb{R}^{H_5} . 
    \end{gathered}
\end{equation}
Then Eqs.~\eqref{eq: temp 1, proof of ua of r-ftnet, firstly step 5} and~\eqref{eq: temp 2, proof of ua of r-ftnet, firstly step 5} imply that 
\begin{equation}
    \label{eq: relation between y and c5 p5}
    \max_{t \in [T]} \left\Vert \bm{y}_t^{(1)} - \mathbf{C}_5 \bm{p}_t^{(5)} \right\Vert_\infty \leqslant \varepsilon_5 . 
\end{equation}
\textbf{Step 6}. The final additive $\mathrm{FTNet}$ is constructed to approximate the target $\mathrm{DODS}$. Let $H = H_3 + H_5$, and define the additive $\mathrm{FTNet}$ $f_{+,\mathrm{R}}$ as follows 
\begin{equation}
    \label{eq: def of additive ftnet, proof of ua of r-ftnet, step 6}
    \mathbf{A} = \left[ 
        \begin{array}{c}
            \mathbf{A}_3 \\ 
            \mathbf{A}_5 
        \end{array}
    \right] ,
    \quad
    \mathbf{B} = \left[ 
        \begin{array}{ccc}
            \mathbf{B}_3 & \bm{0} \\ 
            \mathbf{B}_6 & \bm{0} 
        \end{array}
    \right] , 
    \quad
    \bm{\zeta} = \left[ 
        \begin{array}{c}
            \bm{\theta}_3 \\ 
            \bm{\theta}_5 
        \end{array}
    \right] ,
    \quad
    \mathbf{C} = \left[ 
        \begin{array}{cc}
            \bm{0} & \mathbf{C}_5 
        \end{array}
    \right] , 
    \quad
    \bm{q}_0 = \left[ 
        \begin{array}{c}
            \bm{q}_0^{(3)} \\
            \bm{q}_0^{(5)}
        \end{array}
    \right] , 
\end{equation}
where $\mathbf{B}_6 = [ \mathbf{0} , \mathbf{B}_5 ]$ pads the matrix $\mathbf{B}_5$ with $0$. We claim that $\bm{p}_t = [ \bm{p}_t^{(3)} ; \bm{p}_t^{(5)} ]$ and $\bm{q}_t = [ \bm{q}_t^{(3)} ; \bm{q}_t^{(5)} ]$ hold for any $t \in [T]$. The proof of $\bm{q}_t$ is similar to that of $\bm{p}_t$, and we only prove the claim of $\bm{p}_t$ using mathematical induction as follows. 
\begin{enumerate}
    \item For $t=1$, one has 
    \begin{equation*}
        \begin{aligned}
            \bm{p}_1 
            =& \sigma_1 \left( \mathbf{A} \bm{x}_1 + \mathbf{B} \bm{q}_0 - \bm{\zeta} \right) \\ 
            =& \sigma_1 \left( \left[ 
                \begin{array}{c}
                    \mathbf{A}_3 \bm{x}_1 + \mathbf{B}_3 \bm{q}_0^{(3)} - \bm{\theta}_3 \\ 
                    \mathbf{A}_5 \bm{x}_1 + \left[ 
                        \begin{array}{cc}
                            \bm{0} & \mathbf{B}_5 
                        \end{array}
                    \right] \bm{q}_0^{(3)} - \bm{\theta}_5 
                \end{array}
            \right] \right) \\ 
            =& \left[ \bm{p}_1^{(3)} ; \bm{p}_1^{(5)} \right] , 
        \end{aligned}
    \end{equation*}
    where the first equality holds because of the definition of $\bm{p}_t$ with $t=1$ in Eq.~\eqref{eq: mapping of addftnet, main body}, the second equality holds according to Eq.~\eqref{eq: def of additive ftnet, proof of ua of r-ftnet, step 6}, and the third equality holds based on the definition of $\bm{q}_0^{(3)}$ in Eq.~\eqref{eq: def of additive ftnet, proof of ua of r-ftnet, step 6}, the definitions of $\bm{p}_t^{(3)} , \bm{p}_t^{(5)}$ with $t=1$ in Eqs.~\eqref{eq: addftnet in step 4, in ua of addftnet}~and~\eqref{eq: def of pt5, proof of ua of r-ftnet, firstly step 5}. 
    \item Suppose that the claim holds for $t=k$ where $k \in [T-1]$. Thus, one has 
    \begin{equation*}
        \begin{aligned}
            \bm{p}_{k+1}
            =& \sigma_1 \left( \mathbf{A} \bm{x}_{k+1} + \mathbf{B} \bm{q}_k - \bm{\zeta} \right) \\ 
            =& \sigma_1 \left( \left[ 
                \begin{array}{c}
                    \mathbf{A}_3 \bm{x}_{k+1} + \mathbf{B}_3 \bm{q}_k^{(3)} - \bm{\theta}_3 \\ 
                    \mathbf{A}_5 \bm{x}_{k+1} + \left[ 
                        \begin{array}{cc}
                            \bm{0} & \mathbf{B}_5 
                        \end{array}
                    \right] \bm{q}_k^{(3)} - \bm{\theta}_5 
                \end{array}
            \right] \right) \\ 
            =& \left[ \bm{p}_{k+1}^{(3)} ; \bm{p}_{k+1}^{(5)} \right] , 
        \end{aligned}
    \end{equation*}
    where the first equality holds from the definition of $\bm{p}_t$ with $t=k+1$, the second equality holds because of the definitions of $\mathbf{A} , \mathbf{B} , \bm{\zeta} , \bm{q}_0$, and the third equality holds based on the definitions of $\bm{p}_t^{(3)} , \bm{p}_t^{(5)}$ with $t=k+1$, and the conclusion $\bm{q}_t^{(3)} = [ \times_t ; \bm{q}_t^{(2)} ] $ with $t=k$.  
\end{enumerate}
For any $t \in [T]$, one has 
\begin{equation*}
    \begin{aligned}
        \left\Vert y_t - y_{+,t} \right\Vert_\infty
        &\leqslant \left\Vert \psi(\bm{h}_t) - \psi \left( \mathbf{C}_3 \bm{p}_t^{(3)} \right) \right\Vert_\infty + \left\Vert \psi \left( \mathbf{C}_3 \bm{p}_t^{(3)} \right) - \bm{y}_t^{(1)} \right\Vert_\infty + \left\Vert \bm{y}_t^{(1)} - \mathbf{C}_5 \bm{p}_t^{(5)} \right\Vert_\infty \\ 
        &\leqslant u \left( \left\Vert \bm{h}_t - \mathbf{C}_3 \bm{p}_t^{(3)} \right\Vert_\infty \right) + \varepsilon_4 + \varepsilon_5 \\ 
        &\leqslant u(\varepsilon_3) + \varepsilon_4 + \varepsilon_5 , 
    \end{aligned}
\end{equation*}
where the first inequality holds because of the triangle inequality, the definitions of $\mathrm{DODS}$ and $\hat{\bm{y}}_t$, the second inequality holds based on the definition of $u(a)$, Eq.~\eqref{eq: ua of psi} with $\bm{h} = \mathbf{C}_3 \bm{p}_t^{(3)}$, and Eq.~\eqref{eq: relation between y and c5 p5}, and the third inequality holds in view of Eq.~\eqref{eq: relation between m and c3 p3}. Since $u(a)$ is continuous, and $u(0)=0$, for any $\varepsilon_6 > 0$, there exists $\delta_4(\varepsilon_6) > 0 $, such that for any $\varepsilon_3 \leqslant \delta_4(\varepsilon_6)$, one has $u(\varepsilon_3) \leqslant \varepsilon_6$. Let $ \varepsilon_5 = \varepsilon_4 = \varepsilon_6 = \varepsilon / 3$, then one has $\max_{t \in [T]} \Vert y_t - y_{+,t} \Vert_\infty \leqslant \varepsilon$, i.e., 
\begin{equation*}
    \sup_{ \bm{x}_{1:T} \in K^T } \Vert f_\mathrm{D}(\bm{x}_{1:T}) - f_{+,\mathrm{R}}(\bm{x}_{1:T}) \Vert_\infty \leqslant \varepsilon , 
\end{equation*}
which completes the proof. $\hfill \square$ 

\section{Useful Lemmas}

\begin{lemma}
    \label{lem: ua with row independent matrix}
    Suppose that $\sigma : \mathbb{R} \rightarrow \mathbb{R}$ is continuous almost everywhere and not polynomial almost everywhere. Then for any $\varepsilon > 0$, any continuous function $f : \mathbb{R}^I \rightarrow \mathbb{R}^O$, and any compact set $K \subset \mathbb{R}^I$, there exist $H \in \mathbb{N}^+$, $\mathbf{W} \in \mathbb{R}^{H \times I}$, $\bm{\theta} \in \mathbb{R}^H$, and row independent $\mathbf{U} \in \mathbb{R}^{O \times H}$, such that 
    \begin{equation*}
        \Vert f(\bm{x}) - \mathbf{U} \sigma ( \mathbf{W} \bm{x} - \bm{\theta} ) \Vert_{L^\infty(K)} \leqslant \varepsilon . 
    \end{equation*}
\end{lemma}

\noindent \textbf{Proof.} For any $\varepsilon > 0$, continuous function $f : \mathbb{R}^I \rightarrow \mathbb{R}^O$, and compact set $K \subset \mathbb{R}^I $, Lemma~\ref{lem: ua of fnn} indicates that there exist $H_1 \in \mathbb{N}^+$, $\mathbf{W}_1 \in \mathbb{R}^{H_1 \times I}$, $\bm{\theta}_1 \in \mathbb{R}^{H_1}$, and $\mathbf{U}_1 \in \mathbb{R}^{O \times H_1}$, s.t.
\begin{equation*}
    \left\Vert f(\bm{x}) - \mathbf{U}_1 \sigma \left( \mathbf{W}_1 \bm{x} - \bm{\theta}_1 \right) \right\Vert_{L^\infty(K)} \leqslant \varepsilon . 
\end{equation*}
Define a new $\mathrm{FNN}$ with hidden size $H = H_1 + O$ as follows 
\begin{equation*}
    \begin{aligned}
        \mathbf{W} = \left[ 
            \begin{array}{c}
                \mathbf{W}_1 \\ 
                \bm{0} 
            \end{array}
        \right] , 
        \quad 
        \bm{\theta} = \left[ 
            \begin{array}{c}
                \bm{\theta}_1 \\ 
                \bm{0} 
            \end{array}
        \right] , 
        \quad 
        \mathbf{U} = \left[ 
            \begin{array}{cc}
                \mathbf{U}_1 & \mathbf{I}_O 
            \end{array}
        \right] , 
    \end{aligned}
\end{equation*}
where $\mathbf{I}_O$ is the identity matrix of size $O \times O$. Then it is easy to see that $\mathbf{U}$ is row independent and 
\begin{equation*}
    \Vert f(\bm{x}) - \mathbf{U} \sigma ( \mathbf{W} \bm{x} - \bm{\theta} ) \Vert_{L^\infty(K)}
    = \Vert f(\bm{x}) - \mathbf{U}_1 \sigma ( \mathbf{W}_1 \bm{x} - \bm{\theta}_1 ) \Vert_{L^\infty(K)} 
    \leqslant \varepsilon , 
\end{equation*}
which completes the proof. $\hfill \square$ 

\begin{lemma}
    \label{lem: u is continuous}
    Let $\varphi : \mathbb{R}^n \rightarrow \mathbb{R}$ be continuous, and $K_2 \subset \mathbb{R}^n$ is a convex compact set. Then $u(a) = \sup \{ \Vert \varphi(\bm{y}) - \varphi(\bm{z}) \Vert_\infty \mid \bm{y} , \bm{z} \in K_2 , \Vert \bm{y} - \bm{z} \Vert_\infty \leqslant a \}$ is continuous on $[0,+\infty)$.
\end{lemma}

\noindent \textbf{Proof.} The proof is divided into several steps. \\ 
\textbf{Step 1.} We prove that $u$ is well-defined and bounded. Since any continuous function is bounded on any compact set, there exists $U_\varphi \in \mathbb{R}$, such that $\Vert \varphi(\bm{y}) \Vert_\infty \leqslant U_\varphi $ holds for any $\bm{y} \in K_2$. Then according to the triangle inequality,  $\Vert \varphi(\bm{y}) - \varphi(\bm{z}) \Vert_\infty \leqslant 2 U_\varphi $ holds for any $ \bm{y} , \bm{z} \in K_2$, i.e., $ |u(a)| \leqslant 2 U_\varphi $ holds for any $a \in [0,+\infty)$. Thus, $u(a)$ is well-defined and bounded on $[0,+\infty)$. \\ 
\textbf{Step 2.} It is obvious that $u(0) = 0$. \\ 
\textbf{Step 3.} We prove that $u$ is monotonically increasing. Let $ 0 \leqslant a_1 < a_2 $. For any $\varepsilon > 0 $, according to the definition of supremum, there exist $\bm{y}_1 , \bm{z}_1 \in K_2$, such that $\Vert \varphi(\bm{y}_1) - \varphi(\bm{z}_1) \Vert_\infty \geqslant u(a_1) - \varepsilon$ and $\Vert \bm{y}_1 - \bm{z}_1 \Vert_\infty \leqslant a_1$. Since $ a_1 < a_2 $, one has $ \Vert \bm{y}_1 - \bm{z}_1 \Vert_\infty \leqslant a_2 $. Thus, one has 
\begin{equation*}
    u(a_2) \geqslant \Vert \varphi(\bm{y}_1) - \varphi(\bm{z}_1) \Vert_\infty \geqslant u(a_1) - \varepsilon . 
\end{equation*}
According to the arbitrariness of $\varepsilon$, one has $u(a_2) \geqslant u(a_1)$. Therefore, $u(a)$ is a monotonically increasing function. \\ 
\textbf{Step 4.} We prove that $u$ is right continuous. Let $b \in [0,+\infty)$ be an arbitrary non-negative real number. Since $u(a)$ is bounded and monotonically increasing on $[0,+\infty)$, the limit $\lim_{a \rightarrow b_+} u(a)$ exists. Let $u_+ = \lim_{a \rightarrow b_+} u(a)$ denote this limit. If $ u_+ \neq u(b) $, then one has $ u_+ > u(b) $ since $u(a)$ is monotonically increasing. Since any continuous function on a compact set is uniformly continuous, there exists $\delta > 0$, such that for any $\bm{y} , \bm{z} \in K_2$, $\Vert \bm{y} - \bm{z} \Vert_\infty \leqslant \delta$ indicates $\Vert \varphi(\bm{y}) - \varphi(\bm{z}) \Vert_\infty \leqslant [u_+ - u(b)] / 3$. 
Since $u(a)$ is monotonically increasing, one has $u(b+\delta) \geqslant u_+$. According to the definition of supremum, there exist $\bm{y}_2 , \bm{z}_2 \in K_2$, such that $\Vert \bm{y}_2 - \bm{z}_2 \Vert_\infty \leqslant b + \delta$ and 
\begin{equation*}
    \Vert \varphi(\bm{y}_2) - \varphi(\bm{z}_2) \Vert_\infty \geqslant u_+ - [u_+ - u(b)] / 3 . 
\end{equation*}
Let $\bm{\xi} = \lambda \bm{z}_2 + (1-\lambda) \bm{y}_2$, where $\lambda = b(b+\delta)^{-1} \in [0,1]$. Since $\bm{y}_2 , \bm{z}_2 \in K_2$, and $K_2$ is convex, one has $\bm{\xi} \in K_2$. According to the homogeneity of norm, one has 
\begin{equation*}
    \begin{aligned}
        \Vert \bm{\xi} - \bm{y}_2 \Vert_\infty 
        =&~ \lambda \Vert \bm{z}_2 - \bm{y}_2 \Vert_\infty 
        \leqslant \lambda (b+\delta) 
        = b , \\ 
        \Vert \bm{z}_2 - \bm{\xi} \Vert_\infty 
        =&~ (1-\lambda) \Vert \bm{z}_2 - \bm{y}_2 \Vert_\infty 
        \leqslant (1-\lambda) (b+\delta) 
        = \delta . 
    \end{aligned}
\end{equation*}
Thus, one has 
\begin{equation*}
    \begin{aligned}
        u(b) 
        \geqslant&~ \Vert \varphi(\bm{\xi}) - \varphi(\bm{y}_2) \Vert_\infty \\ 
        \geqslant&~ \Vert \varphi(\bm{z}_2) - \varphi(\bm{y}_2) \Vert_\infty - \Vert \varphi(\bm{z}_2) - \varphi(\bm{\xi}) \Vert_\infty \\ 
        \geqslant&~ \left( u_+ - [u_+-u(b)] / 3 \right) - [u_+-u(b)] / 3 \\ 
        =&~ u(b) + [u_+-u(b)] / 3 \\ 
        >&~ u(b) , 
    \end{aligned}
\end{equation*}
where the first inequality holds from $ \Vert \bm{\xi} - \bm{y}_2 \Vert_\infty  = b $, the second inequality holds based on the triangle inequality, and the third inequality holds because of the definitions of $\bm{y}_2 , \bm{z}_2$, and $ \Vert \bm{z}_2 - \bm{\xi} \Vert_\infty = b $. The above inequality is a contradiction, which means that $ u_+ \neq u(b) $ does not hold. Therefore, one has $ u_+ = u(b) $, which means that $u(a)$ is right continuous. \\ 
\textbf{Step 5.} Similarly, we can prove that $u(a)$ is left continuous. Therefore, $u(a)$ is continuous. $\hfill \square$ 

\begin{lemma}
    \label{lem: limit is 0}
    Let $a_0 = 0$. For any $t \in [T]$, let $a_t = u(a_{t-1}) + \varepsilon$, where $T \in \mathbb{N}^+$ is a positive integer, $u : \mathbb{R} \rightarrow \mathbb{R}$ is continuous, and $u(0) = 0$. Then $\lim_{\varepsilon \rightarrow 0+} a_t = 0$ holds for any $t \in [T]$. 
\end{lemma}

\noindent \textbf{Proof.} We prove this lemma by mathematical induction. 
\begin{enumerate}
    \item For $t=1$, one has 
    \begin{equation*}
        \lim_{\varepsilon_1 \rightarrow 0_+} a_1 
        = \lim_{\varepsilon_1 \rightarrow 0_+} u(a_0) + \varepsilon_1 
        = 0 + 0 
        = 0 ,
    \end{equation*}
    where the first equality holds from the definition of $a_t$ with $t=1$, and the second equality holds based on $a_0=0$ and $u(0)=0$.
    Thus, the conclusion holds for $t=1$. 
    \item If the conclusion holds for $t=k$ where $k \in [T-1]$, then 
    \begin{equation*}
        \begin{aligned}
            \lim_{\varepsilon_1 \rightarrow 0_+} a_{k+1} 
            = \lim_{\varepsilon_1 \rightarrow 0_+} u(a_k) + \varepsilon_1
            = u \left( \lim_{\varepsilon_1 \rightarrow 0_+} a_k \right) + 0
            = u(0) + 0 
            = 0 ,
        \end{aligned}
    \end{equation*}
    where the first equality holds from the definition of $a_t$ with $t=k+1$, the second equality holds based on the continuity of $u(a)$, the third equality holds because of the induction hypothesis, and the fourth equality holds since $u(0)=0$.
    Thus, the conclusion holds for $t=k+1$. 
\end{enumerate}
Then mathematical induction completes the proof. $\hfill \square$

\bibliography{mybibfile}\bibliographystyle{plain}

\end{document}